\documentclass{article}

\PassOptionsToPackage{numbers, compress}{natbib}

\usepackage[preprint]{neurips_2023}

\usepackage{floatrow}
\usepackage{wrapfig}
\usepackage[utf8]{inputenc} % allow utf-8 input
\usepackage[T1]{fontenc}    % use 8-bit T1 fonts
\usepackage{hyperref}       % hyperlinks
\usepackage{url}            % simple URL typesetting
\usepackage{booktabs}       % professional-quality tables
\usepackage{amsfonts}       % blackboard math symbols
\usepackage{nicefrac}       % compact symbols for 1/2, etc.
\usepackage{microtype}      % microtypography
\usepackage{amsmath}      % microtypography
\usepackage{xcolor}         % colors
\usepackage{graphicx}         % colors
\usepackage{todonotes}         % colors
\usepackage{algorithm}
\usepackage{floatrow}
\usepackage{algpseudocode}
\usepackage{float}
\usepackage{enumitem}
\usepackage{subcaption}

\algdef{SE}[SUBALG]{Indent}{EndIndent}{}{\algorithmicend\ }%
\algtext*{Indent}
\algtext*{EndIndent}

\usepackage{amsmath,amsfonts,bm}

\newcommand{\ind}{\perp\!\!\!\!\perp}

\def\eqref#1{equation~\ref{#1}}
\def\1{\bm{1}}

\DeclareMathAlphabet{\mathsfit}{\encodingdefault}{\sfdefault}{m}{sl}
\SetMathAlphabet{\mathsfit}{bold}{\encodingdefault}{\sfdefault}{bx}{n}

\def\sN{{\mathbb{N}}}

\title{Enhancing Representation Learning on High-Dimensional, Small-Size Tabular Data: \\ A Divide and Conquer Method with Ensembled VAEs}

\author{%
  Navindu Leelarathna \\
  University of Cambridge\\
  \texttt{nyl25@cam.ac.uk} \\
  \And
  Andrei Margeloiu \\
  University of Cambridge\\
  \texttt{am2770@cam.ac.uk} \\
  \And
  Mateja Jamnik \\
  University of Cambridge\\
  \texttt{mj201@cam.ac.uk} \\
  \And
  Nikola Simidjievski \\
  University of Cambridge\\
  \texttt{ns779@cam.ac.uk} \\
}

\begin{document}
\maketitle

\looseness=-1
\begin{abstract}
  Variational Autoencoders and their many variants have displayed impressive ability to perform dimensionality reduction, often achieving state-of-the-art performance. Many current methods however, struggle to learn good representations in High Dimensional, Low Sample Size (HDLSS) tasks, which is an inherently challenging setting. We address this challenge by using an ensemble of lightweight VAEs to learn posteriors over subsets of the feature-space, which get aggregated into a joint posterior in a novel divide-and-conquer approach. Specifically, we present an alternative factorisation of the joint posterior that induces a form of implicit data augmentation that yields greater sample efficiency. Through a series of experiments on eight real-world datasets, we show that our method learns better latent representations in HDLSS settings, which leads to higher accuracy in a downstream classification task. Furthermore, we verify that our approach has a positive effect on disentanglement and achieves a lower estimated Total Correlation on learnt representations. Finally, we show that our approach is robust to partial features at inference, exhibiting little performance degradation even with most features missing.
  
\end{abstract}

\section{Introduction}
% SET THE STA OF HDLSS
High Dimensional, Low Sample Size (HDLSS) data occurs ubiquitously in clinical research and gene expression analysis but is challenging to work with for two reasons. Firstly, as a consequence of high-dimensionality, raw data often only sparsely covers the input space, an issue commonly referred to as the \textit{curse of dimensionality}. Secondly, analysing such data is often computationally challenging due to the size of models required and vast quantities of training data needed. Dimensionality reduction has emerged as a hugely popular technique to address this issue and is often used as a preprocessing step to transform high-dimensional data into a lower-dimensional space (whilst retaining meaningful information) that is more readily suitable for downstream tasks. 

% Limitations with previous work
In particular, VAEs~\cite{kingma_auto-encoding_2013} and their variants have emerged as a dominant architecture for dimensionality reduction and have displayed an impressive performance in recent years. However, the curse of dimensionality remains, and the majority of current approaches still require large volumes of input data to learn high-quality latent representations. As a result, many state-of-the-art models struggle with HDLSS data and either overfit or fail to learn high-quality representations, which we attribute to overparameterisation and low sample-efficiency.  

% our hypothesis, and argue why it is sensible
% provide motivation for THE METHOD; why do you do this?
% include divide-and-conquer

To this end, we propose a novel approach to representation learning for tabular data, Ensemble-VAE (EnVAE), which trains an ensemble of lightweight VAEs on smaller subsets of features. We hypothesise that using a divide-and-conquer approach to perform dimensionality reduction is more tractable since (i) learning good global representations from subsets of features will lead to latent representations that generalise better to new data and (ii) reducing the dimensionality / samples ratio intuitively decreases the complexity of each instance. Concretely, our method artificially splits the features into pseudo-groups and learns ``sub-latents'' which are eventually aggregated using a Mixture-of-Product-of-Experts~\cite{sutter_generalized_2021} function. We argue that this formulation induces implicit data augmentation, improving sample efficiency and robustness to overfitting, resulting in higher-quality latent representations being learnt. We validate our hypothesis by testing our proposed model against standard baselines on eight different HDLSS datasets. 
We also note that our approach is simple to implement and introduces minimal computational overhead over a standard VAE.
Our contributions \footnote{We have included the code and will share it publicly after publication.} are summarised as:
%
% @Nav, can you add 2-3 more sentences on why the D-and-Q is a sensible approach, esp on HDLSS? Why do you want to have an ensemble of VAEs? Why smaller VAEs? Can you make any argument wrt feature bagging from RF?  
% VAE's\cite{wu_multimodal_2018,shi_variational_2019,sutter_generalized_2021}. Whilst originally designed for aggregating multimodal data, we show that their benefits extend to single-modality tabular data as well. 
% \Andrei{Update contributions at the end.}
\begin{enumerate}[topsep=0pt,itemsep=3pt]
    \item We propose a novel method EnVAE that uses an ensemble of VAEs to learn higher-quality latent representations unsupervised in HDLSS settings (Section~\ref{sec:method}). Our approach is task-agnostic, support missing data at both training and inference and yields substantial performance gains, outperforming other popular methods in all HDLSS tasks we looked at.
    \item We demonstrate that our approach learns better latent representations in eight real-world HDLLS datasets compared to benchmark models even as we further constrain the number of training samples (Section~\ref{section:accuracy-downstream}).
    \item Via post-hoc analysis, we show quantitatively that EnVAE learns representations that achieve greater disentanglement compared to a standard $\beta$-VAE by estimating Total Correlation of the learnt latent representations (Section~\ref{section:dissecting-model}).
    \item We demonstrate that our approach is robust to missing features at inference time, producing high-quality latent representations that are usable for downstream classification tasks even when most of the features are missing (Section~\ref{section:dissecting-model}).
\end{enumerate}

\section{Background}

\label{section:background}

% OVERVIEW SECTION
%This section provides the necessary background to understand our proposed method, presented in the next section. Once we have explained our method, we will connect it to related work in Section~\ref{section:related-work}.

% PROBLEM: APPROXIMATE THE CONDITIONAL DISTRIBUTION
\looseness-1
\textbf{Problem definition}. Let $\mathcal{X}=\{\pmb{x}^{(i)}\}_{i=1,...,N}$ be a set of $N$ samples $\pmb{x}^{(i)} \in \mathbb{R}^D$, each having $D$ features. We generally refer to a sample as $\pmb{x}$. We assume a simple latent variable model in which samples $\pmb{x}$ are generated from an unknown conditional distribution $p(\pmb{x}|\pmb{z})$ on the latent variable $\pmb{z} \sim p(\pmb{z})$. We consider the marginal distribution $p(\pmb{z})$ of the latent variable to be known, and we seek to learn the conditional 
distribution $p(\pmb{x}|\pmb{z})$.

% WHAT VAEs TRY TO DO
\textbf{Variational Autoencoders} (VAEs)~\cite{kingma_auto-encoding_2013} are a method for learning the conditional distribution $p(\pmb{x}|\pmb{z})$. They consist of two components, both of which are implemented as neural networks: (i) an encoder $q_{\phi}(\pmb{z}|\pmb{x}): X \rightarrow Z$ with parameters $\phi$ which takes as input a sample $\pmb{x}$ and outputs the parameters of the conditional distribution $p(\pmb{z} | \pmb{x})$, and (ii) a decoder $p_{\theta}(\pmb{x}|\pmb{z}): Z \rightarrow \hat{X}$ with parameters $\theta$ which takes as input the latent representation $\pmb{z}$ and outputs an approximated (usually called ``reconstructed'') version of the input $\hat{\pmb{x}} \approx \pmb{x}$.

% VAE OPTIMISATION
% because the true posterior density $p_{\theta}(\pmb{z}|\pmb{x})$ is itself computationally intractable
Training VAEs require maximising the marginal likelihood of the data $p(\pmb{x})=\int p(\pmb{z}) p(\pmb{x}|\pmb{z}) \,dz$. However, directly optimising the marginal likelihood is generally intractable. Instead, VAEs are trained using variational inference. Thus, a learnable variational distribution $q_{\phi}(\pmb{z}|\pmb{x})$ (which is learned by the VAE's encoder) is introduced to approximate the true posterior $p(\pmb{z}|\pmb{x})$. The encoder and the decoder are jointly optimised by maximising a lower bound of the log-likelihood, called the Evidence Lower BOund (ELBO)~\citep{kingma_auto-encoding_2013}
\begin{equation}
     \text{ELBO}(\pmb{x}) := \mathbb{E}_{\pmb{z} \sim q_\phi(\pmb{z}|\pmb{x})} \left[ \log p_{\theta}(\pmb{x}|\pmb{z}) \right] - \mathbb{KL}( q_\phi(\pmb{z}|\pmb{x})|| p(\pmb{z})) \leq \log p(\pmb{x})
\end{equation}
The first term is a reconstruction loss of the data from the approximated latent representation. The second term regularises the approximated posterior to be similar to the prior distribution $p(\pmb{z})$. In the case of a standard VAE, the prior $p(\pmb{z}) = \mathbb{N}(0,I)$ is an isotropic Gaussian distribution. $\beta$-VAEs~\citep{higgins_irina_beta-vae_2017} generalise VAEs by adding a hyperparameter $\beta$ in the ELBO which balances the trade-off between the two loss terms.

\section{Ensembled VAEs} 
\label{sec:method}

\begin{figure}[t!]
    \centering
    \includegraphics[width=\linewidth, clip, trim=15 0 45 10 10]{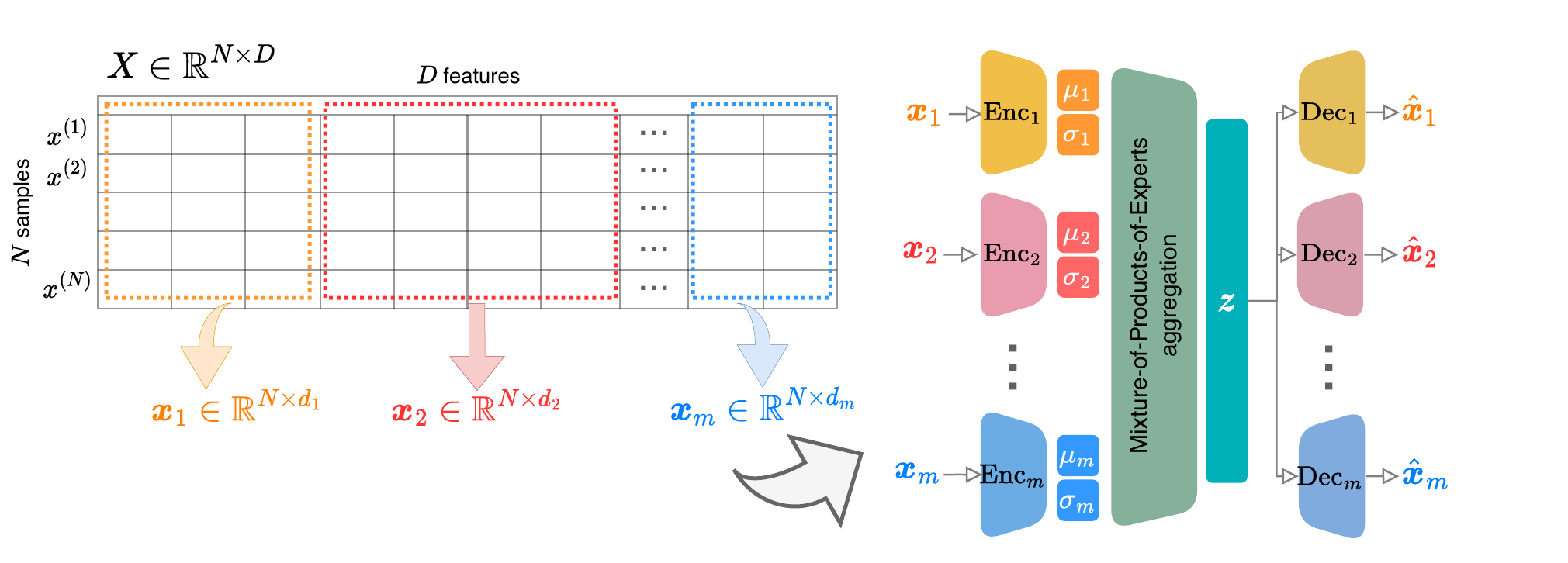}
    \caption{Architecture for our proposed method, \textbf{EnVAE}. (A) Split the data into $m$ feature groups $\pmb{x}_1, \pmb{x}_2, ..., \pmb{x}_m$. (B) Pass each group $\pmb{x}_i$ through a dedicated encoder $\text{Enc}_i$ to compute each group's posterior distribution (specified by a Gaussian distribution $q_\phi(\pmb{z}|\pmb{x}_i) \sim \mathcal{N}(\pmb{\mu}_i, \pmb{\sigma}_i)$) (C) Aggregate the latent representations into a global representation, which is then passed to each of the $m$ decoders, $\text{Dec}_i$. \looseness=-1}
    \label{fig:model_architecture}
\end{figure}

% OVERVIEW HIGH-LEVEL IDEA
Our method is a divide-and-conquer strategy for constructing and training VAEs on HDLSS tabular data. Instead of training a single model on all $D$ features (as standard VAEs do), we propose  to train an ensemble of lightweight VAEs on subsets of the feature-space. Figure~\ref{fig:model_architecture} depicts our approach which we name \textbf{En}semble-\textbf{VAE} (\textbf{EnVAE}).

% NOTATION OF m GROUPS
Specifically, we split the $D$ features into $m$ disjoint groups -- thus we can view training samples $\pmb{x}$ as a concatenation of $m$ groups of features $\pmb{x}=[\pmb{x}_1 ,... , \pmb{x}_m ]$ where the number of features in $\pmb{x}_i$ is $d_i$. We assume the groups of features $\pmb{x}_j$ are conditionally independent on the latent variable $\pmb{z}$ such that:

\vspace{-12pt}
\begin{align}
    \label{eq:independence}
    p_\theta(\pmb{x} | \pmb{z}) &= p_{\theta}(\pmb{x}_1,\pmb{x}_2,..., \pmb{x}_m | \pmb{z}) \nonumber\\
    &= p_{\theta}(\pmb{x}_1|\pmb{z})\cdot p_{\theta}(\pmb{x}_2|\pmb{z}) \cdot.... \cdot p_{\theta}(\pmb{x}_m|\pmb{z}) && \text{we assume $ \pmb{x}_1, \pmb{x}_2, ..., \pmb{x}_m \ind \pmb{z}$}
\end{align}
Under this factorisation, the ELBO becomes:
\begin{equation}
    \text{ELBO}(\pmb{x}) = \mathbb{E}_{z \sim q_\phi(z|x)} \left[ \sum_{i=1}^m \log p_{\theta}(\pmb{x}_i|\pmb{z}) \right] - \mathbb{KL}( q_\phi(\pmb{z}|\pmb{x})|| p_\theta(\pmb{z})) 
    \label{eq:multimodal_elbo}
\end{equation}

Naively, computing the posterior $q_{\phi}(\pmb{z}|\pmb{x}):=q_\phi (\pmb{z}| \pmb{x}_1, \pmb{x}_2,...,\pmb{x}_m)$ in Equation~\ref{eq:multimodal_elbo} requires all feature groups $\pmb{x}_1, \pmb{x}_2,...,\pmb{x}_m$. To avoid this, we decompose the posterior $q_\phi (\pmb{z}| \pmb{x})$ as a Mixture-of-Product-of-Experts (MoPoE)~\citep{sutter_generalized_2021}. In terms of notation, let $\sN_m = \{1,2,...,m\}$ represent the set of feature groups' indices. Its power set is given by $\mathcal{P}(\sN_m)$. Any set of feature groups is denoted as $A \in \mathcal{P}(\mathbb{N}_m)$. The feature space defined by a set of feature groups is represented as $\pmb{x}_A := \{\pmb{x}_i : i \in A\}$.

\textbf{Joint posterior as a Mixture of Experts.} We define the joint posterior $q_{\phi}(\pmb{z} | \pmb{x})$ as a mixture of all posteriors $q_{\phi}(\pmb{z}| \pmb{x}_A)$ from sets $A \in \mathcal{P}(\mathbb{N}_m)$ of feature groups:
\begin{align}
    \label{eq:mopoe}
    q_{\phi}(\pmb{z} | \pmb{x}) = \frac{1}{|\mathcal{P}(\mathbb{N}_m)|} \sum_{A \in \mathcal{P}(\mathbb{N}_m)}   q_{\phi}(\pmb{z} | \pmb{x}_A) && \text{the approximate posterior as a mixture}
\end{align}

\textbf{Set posterior as a Product of Experts.} The posterior $q_{\phi}(\pmb{z}| \pmb{x}_A)$ given a set $A$ of feature groups is decomposed as a Product-of-Experts:
\begin{align}
    \label{eq:poe}
    q_{\phi}(\pmb{z}| \pmb{x}_A) &\propto p_{\theta}(\pmb{x}_A | \pmb{z}) p(\pmb{z}) \nonumber && \text{Bayes rule} \\
     &= \prod_{i \in A} p_{\theta}(\pmb{x}_i | \pmb{z}) \times p(\pmb{z}) && \text{we assume $ \pmb{x}_1, \pmb{x}_2, ..., \pmb{x}_m \ind \pmb{z}$ (see Eq. \ref{eq:independence})} \nonumber \\
    &= \prod_{i \in A} \left[ \frac{q_{\phi}(\pmb{z}|\pmb{x}_i) p(\pmb{x}_i)}{p(\pmb{z})} \right] \times p(\pmb{z}) \nonumber && \text{Bayes rule}\\
    &= \frac{  \prod_{i \in A}  q_{\phi}(\pmb{z}|\pmb{x}_i)   }{  \prod_{j=1}^{|A|-1}  p(\pmb{z})  }
\end{align}

Hence the posterior $q_{\phi}(\pmb{z}| \pmb{x}_A)$ from a set $A$ of feature groups is the product of the posteriors $q_{\phi}(\pmb{z}|\pmb{x}_i)$ of each feature group $i \in \{\pmb{x}_i : i \in A\}$. By letting the underlying inference networks compute $\hat{q}_{\phi}(\pmb{z}|\pmb{x}_i) = \frac{q_{\phi}(\pmb{z}|\pmb{x}_i) }{p(\pmb{z})}$, Equation~\ref{eq:poe}  simplifies to:
\begin{equation}
    q_{\phi}(\pmb{z}| \pmb{x}_A) \propto p(\pmb{z}) \prod_{i \in A}\hat{q}_{\phi}(\pmb{z}|\pmb{x}_i)
\end{equation}
\textbf{Our method performs implicit data augmentation.} We argue that our decomposition of the posterior results in an implicit form of data augmentation, which should aid generalisation, especially on low sample size datasets. Analysing the MoPoE approximation for the posterior in Equation~\ref{eq:mopoe}, we observe that we compute a joint posterior for all possible subsets of groups, hence our objective has $2^m-1$ terms. Intuitively, this amounts to using many different subsets $\pmb{x}_A$ of features to reconstruct the complete sample $\hat{\pmb{x}}$, improving sample efficiency and effectively acting as a form of data augmentation. This process is akin to reconstructing an image from patches \citep{he_masked_2022}. By substituting Equation~\ref{eq:mopoe} into the ELBO expression from Equation~\ref{eq:multimodal_elbo} we can reinterpret the objective as:
\begin{equation}
    \text{ELBO}(\pmb{x}) = \frac{1}{|\mathcal{P}(\mathbb{N}_m)|} \underbrace{\sum_{A \in \mathcal{P}(\mathbb{N}_m)} \mathbb{E}_{\pmb{z} \sim q_\phi(\pmb{z}|\pmb{x}_A)} \left[ \sum_{i=1}^m \log p_{\theta}(\pmb{x}_i|\pmb{z}) \right]}_\text{Implicit data augmentation: reconstruct $\pmb{x}$ from every feature set $A$} - \mathbb{KL}( q_\phi(\pmb{z}|\pmb{x})|| p_\theta(\pmb{z})) 
\end{equation}
This factorisation reveals that the total reconstruction loss is a sum of reconstruction losses where the entire set of features $\pmb{x}$ is reconstructed from every possible subset $\pmb{x}_A$ of feature groups. We hypothesise this implicit data augmentation mechanism results in learning posteriors for each group, that when aggregated together generalise better, particularly in low-sample size settings -- and we investigate this empirically in Section~\ref{sec:experiments}.

\section{Experiments}
\label{sec:experiments}
We assess the performance and capabilities of EnVAE by addressing the following aspects:
\begin{itemize}[topsep=0pt, leftmargin=18pt, itemsep=3pt]
\item \textbf{Quality of the latent representations.} We initiate the analysis with a quantitative assessment of (i) the accuracy of downstream classifiers trained using these representations, and (ii) the level of disentanglement within the representations themselves (Section~\ref{section:accuracy-downstream}). Following these evaluations, we employ the learned representations to facilitate data exploration (Section~\ref{section:umap-plots}).
\item \textbf{Examining the model's robustness under varying conditions.} Firstly, we scrutinise the performance of EnVAE when it is trained on extremely limited data sets (Section~\ref{section:ablation-training-size}). Next, we investigate the impact of varying the number of experts within EnVAE, and assess the effects of absent feature groups on the model's performance (Section~\ref{section:dissecting-model}).
\end{itemize}

\looseness-1
\textbf{Datasets.} We concentrate on the HDLSS scenario, examining eight real-world biomedical tabular datasets with 3,312 to 22,293 features. Seven of these datasets encompass 100 to 200 samples, except `meta-pam', which contains nearly 2,000 samples. For more details, refer to Appendix~\ref{appendix:datasets}.

\textbf{Models.} We evaluate EnVAE against $\beta$-VAE~\citep{higgins_irina_beta-vae_2017} (which generalises a standard VAE~\citep{kingma_auto-encoding_2013}), Concrete Autoencoders (CAE)~\citep{balin_concrete_2019} and SubTab~\citep{ucar_subtab_2021}. The baseline model used in all experiments is a standard VAE with the encoder and decoder having 2 hidden layers of size 128. We use ReLU activations followed by batch normalisation layers and dropout layers with $p=0.5$. All ensemble-based VAEs use the same encoders and decoders with 2 hidden layers such that the total number of weights in each model is approximately the same as the benchmark VAE.

\textbf{Training.} We train EnVAE and all benchmark models using five-fold cross-validation across all datasets. Throughout our experiments, we strive for a fair comparison by keeping the optimiser, batch size and common hyperparameters constant across all VAE architectures. We train using a batch size of $32$ and optimise using Adam~\citep{kingma_adam_2014} with standard parameters ($\beta_1$ = 0.9, $\beta_2$ = 0.999) and with learning rate of $0.001$. We use early stopping to terminate training when the validation loss stops improving for 100 epochs. We use a $72/8/20$ test/valid/test split. We perform a grid-search hyperparameter tuning for each method and select optimal settings based on the validation balanced accuracy. For further details on training hyperparameters, see Appendix~\ref{app:hyperparameters}.

We \textbf{evaluate the quality of the learned latent representations} in two ways:
\begin{itemize}[topsep=0pt, leftmargin=18pt, itemsep=3pt]
    \item By assessing the downstream classification accuracy of a classifier trained on the extracted latent representations. After the VAE models have been trained, we compute the latent representations for the entire dataset. Then, we train an MLP classifier five times with different seeds for each VAE model -- while maintaining the same dataset splits -- and resulting in 25 total runs. We report the mean~$\pm$~std of the test balanced accuracy averaged across all 25 runs.
    \item By quantitatively investigating the disentanglement of the representations using Total Correlation~(TC)~\citep{kim_disentangling_2018,chen_isolating_2018}, which is commonly used in representation learning~\citep{locatello_sober_2020} and has good properties~\cite{watanabe_information_1960} capturing the intuition that a disentangled representation should separate the factors of variations~\cite{bengio_representation_2013}.
\end{itemize}

\subsection{Accuracy on Classification Tasks}
\label{section:accuracy-downstream}

\begin{table*}[b!]
\caption{Evaluating the downstream performance of an MLP classifier trained on the latent representations produced by our method EnVAE (with optimal number of experts k) and three benchmark models on the \textit{complete training sets}. We report the mean~$\pm$~std of the test balanced accuracy averaged over 25 runs. Our method, EnVAE, consistently produces better latent representations, providing improvements by up to $11\%$.}
\label{table:downstream-accuracy-full-datasets}
\centering
% \small
\begin{tabular}{lrrrrr}
    \toprule    
Dataset &  EnVAE (ours) & $\beta$-VAE & CAE & SubTab   \\\midrule
meta-pam   &   \pmb{$90.97 \pm2.29$} (k=8) & 83.14 $\pm$2.86 & $ 73.54\pm2.14 $  &  $ 88.11\pm1.51 $           \\
lung & \pmb{$ 94.67\pm6.67 $} (k=8)   & $ 91.52\pm5.23 $  & $ 74.31\pm6.48 $  &  $ 89.12\pm7.29 $           \\
cll  &  \pmb{$ 68.06\pm12.61 $} (k=2) & $ 56.64\pm14.07 $  &  54.35 $\pm$9.10 &$ 54.32\pm11.16 $       \\
breast&   \pmb{$ 86.05\pm12.87 $} (k=2) &  75.59 $\pm$10.32&  $ 78.00\pm7.08 $& $ 65.47\pm16.15 $           \\
mpm & \pmb{$ 95.21\pm4.38 $}  (k=6)&  $ 80.53\pm9.96 $ &  $ 79.33\pm11.69 $& $ 89.08\pm8.83 $            \\
toxicity  &\pmb{$ 74.75\pm8.43 $}  (k=6)&  $ 50.83\pm9.31 $  & 52.35 $\pm$10.85 &  $ 74.16\pm8.23 $    \\
smk & \pmb{$ 67.15\pm8.14 $}  (k=8)&  $ 63.20\pm10.23 $ &54.37 $\pm$7.82  & $ 62.14\pm5.68 $ \\
prostate & \pmb{$ 81.27\pm6.87 $}  (k=8)&  $ 73.16\pm9.86 $  &  56.49 $\pm$8.27&  66.34 $\pm$14.47          \\\bottomrule
\end{tabular}
\end{table*}
% \textbf{Setup.} To evaluate the quality of the latent representation learnt by each model, we first train each model without labels until convergence. Next, the entire dataset is transformed using the learnt encoding that is then used to train a simple MLP classifier network in a supervised manner to perform a relevant classification task. Train-valid-test splits are kept the same in both stages. Whilst training both the latent model and classifier jointly in an end-to-end manner is possible and usually results in better accuracy with regards to the classification task, we choose to split up the training in order to disentangle the impact of the latent model makes and to simulate real-world use-cases. Due to the highly-stochastic nature of training on low-sample sizes, we run each configuration 25 times, using 5-fold cross-validation with 5 different train-validation splits. 

%\textcolor{red}{\textbf{Lorem Ipsum} is simply dummy text of the printing and typesetting industry. Lorem Ipsum has been the industry's standard dummy text ever since the 1500s, when an unknown printer took a galley of type and scrambled it to make a type specimen book. It has survived not only five centuries, but also the leap into electronic typesetting, remaining essentially unchanged. It was popularised in the 1960s with the release of Letraset sheets containing Lorem Ipsum passages, and more recently with desktop publishing software like Aldus PageMaker including versions of Lorem Ipsum.}

\looseness-1
Our central hypothesis is that EnVAE can produce representations that yield stable and improved predictive performance. We evaluate the accuracy of such representations in two scenarios:

\looseness-1
\textbf{Unsupervised setting.} We trained an MLP classification model (as we elaborated earlier) using the latent representations computed by EnVAE and other benchmark models. As indicated by the results Table~\ref{table:downstream-accuracy-full-datasets}, EnVAE consistently surpasses all benchmark models across eight real-world datasets, often by a considerable margin. The most significant improvements are observed in tasks where models that encode all features simultaneously, such as $\beta$-VAE and CAE, struggle the most. Specifically, on datasets `cll', `toxicity', `smk', and `prostate', the improvements range between 10-25\%.

% \end{table*}

Despite also considering groups of features, SubTab lags behind EnVAE. We attribute SubTab's inferior performance to its rigid aggregation mechanism, which applies a convolution on the latent representations from each feature group. These results support that our proposed MoPoE aggregation is better suited to handling HDLSS tasks.

\begin{wrapfigure}{r}{0.57\textwidth}
\vspace{0pt}
\captionof{table}{Performance in a \textit{supervised setting} of EnVAE trained with an additional classification head. We report the mean~$\pm$~std test balanced accuracy averaged over 25 runs. Our method, EnVAE, consistently outperforms an MLP, and it ranks higher than $\beta$-VAE.}
\label{tbl:supervised}
\centering
% \small
\resizebox{\textwidth}{!}{%
\begin{tabular}{lrrr}
    \toprule    
Dataset &EnVAE (ours) & $\beta$-VAE  & MLP  \\
    \midrule
meta-pam   &  $\pmb{89.52\pm2.27}$ &  $85.89 \pm 4.09$&  $87.91 \pm 4.03$      \\
lung   &  $94.30\pm7.03$   &    $\pmb{95.97 \pm 7.07}$&   $83.08 \pm 9.24$    \\
cll  & $\pmb{82.04\pm5.74}$  &    $81.49 \pm 6.92$    &  $74.44\pm9.66$      \\
breast & $\pmb{98.06\pm4.35}$  &  $\pmb{98.06 \pm 4.34}$&  $97.76 \pm 3.72$     \\
mpm   &  $\pmb{99.03\pm0.88}$   & $95.97 \pm 7.07$&  $98.59 \pm 2.89$     \\
toxicity   &  $91.43\pm5.71$   &    $\pmb{93.67 \pm 2.53}$&  $91.71 \pm 3.97$   \\
smk   &  $\pmb{73.72\pm5.50}$    &    $67.67 \pm 5.39$&  $70.11 \pm 8.39$   \\
prostate   &  $\pmb{92.34\pm5.71} $  &    $88.31 \pm 7.04$& $85.68 \pm 13.62$ \\  \midrule
Average rank & \textbf{1.44} & 2.06 & 2.50\\\bottomrule
\end{tabular}%
}
\end{wrapfigure}

\looseness-1
\textbf{Supervised setting.} To further analyse the predictive capabilities of EnVAE, we benchmark our approach in a supervised setting (where the labels are available), by training an end-to-end model that combines EnVAEs with an additional MLP classification head to output the predicted labels. As depicted in Table~\ref{tbl:supervised}, in such a scenario, EnVAE is able to consistently outperform an MLP (with the same architecture as the classification head, but trained on raw data), achieving higher accuracy on  7 out of the 8 tasks, with improvements peaking at 11\%. Compared with the $\beta-$VAE (trained in a similar setting, with an optimal parameter $\beta$), the EnVAE performs competitively, but generally better, exceeding the $\beta-$VAE accuracy on several tasks. % and exceeding the $\beta-$VAE accuracy by up 3-5\% on three tasks.

\subsubsection{Decreasing the Size of the Training Set}
\label{section:ablation-training-size}

\begin{table*}[b!]
\caption{Evaluating the performance of EnVAE when very few training samples are available. We evaluate the downstream performance of a classifier trained on latent representations learnt by EnVAE and three benchmark models where the number of training samples was artificially constrained. For fair comparison, we keep the test sets the same and only reduce the size of the train and validation sets. Notably, our method continues to produce better latent representations over other benchmarks.}
\centering
\small
\begin{tabular}{lrrrrr}
    \toprule    
Dataset & Train Samples&EnVAE (ours) & $\beta$-VAE & CAE & SubTab  \\
    \midrule
meta-pam   &58 & \pmb{$ 81.85\pm2.89 $} & $ 77.49\pm3.70 $ & $ 70.25\pm4.49 $ & $ 61.45\pm2.24 $\\
lung   & 29& \pmb{$ 85.13\pm6.16 $} & $ 78.95\pm8.38 $ & $ 62.33\pm16.39 $ & $ 78.05\pm10.20 $\\
cll   & 33& \pmb{$ 67.64\pm10.49 $} & $ 53.75\pm10.35 $ & $ 52.65\pm11.43 $ & $ 58.04\pm12.88 $\\
breast   &31 & \pmb{$ 83.41\pm9.80 $} & $ 72.16\pm18.64 $ & $ 63.96\pm11.81 $ & $ 66.69\pm17.10 $\\
mpm   & 27& \pmb{$ 90.76\pm7.69 $} & $ 77.99\pm14.03 $ & $ 60.56\pm17.25 $ & $ 80.99\pm15.91 $\\
toxicity    &51 & \pmb{$ 67.06\pm10.09 $} & $ 42.44\pm7.51 $ & $ 48.21\pm10.03 $ & $ 59.99\pm14.12 $\\
smk    & 27& $ 60.15\pm8.28 $ & $ 58.18\pm10.24 $ & $ 51.14\pm7.33 $ & \pmb{$ 61.07\pm7.97 $}\\
prostate    & 15& \pmb{$ 75.52\pm15.47 $} & $ 71.00\pm12.42 $ & $ 59.53\pm9.89 $ & $ 59.25\pm10.22 $\\\bottomrule
\end{tabular}
  \label{table:downstream-accuracy-low-datasets}
\end{table*}

\looseness-1
The ELBO derivation of EnVAE revealed that our MoPoE aggregation performs implicit data augmentation, a technique frequently employed when training with smaller datasets. Consequently, we subject EnVAE to a `stress test' to assess its ability to cope with minimal training sets by restricting the volume of training data. We evaluated the downstream classification accuracy by training an MLP on the latent representations computed by different methods trained on smaller versions of the dataset containing between 15-58 (see Appendix~\ref{appendix:classification_performance} for complete details). The results from Table~\ref{table:downstream-accuracy-low-datasets} show that EnVAE consistently outperforms all models, even on training datasets containing just 15 samples -- which supports our hypothesis that EnVAE performs well, especially on very small datasets.

\begin{figure}[h!]
\begin{subfigure}[b]{0.49\linewidth}
         \centering
         \includegraphics[width=\textwidth]{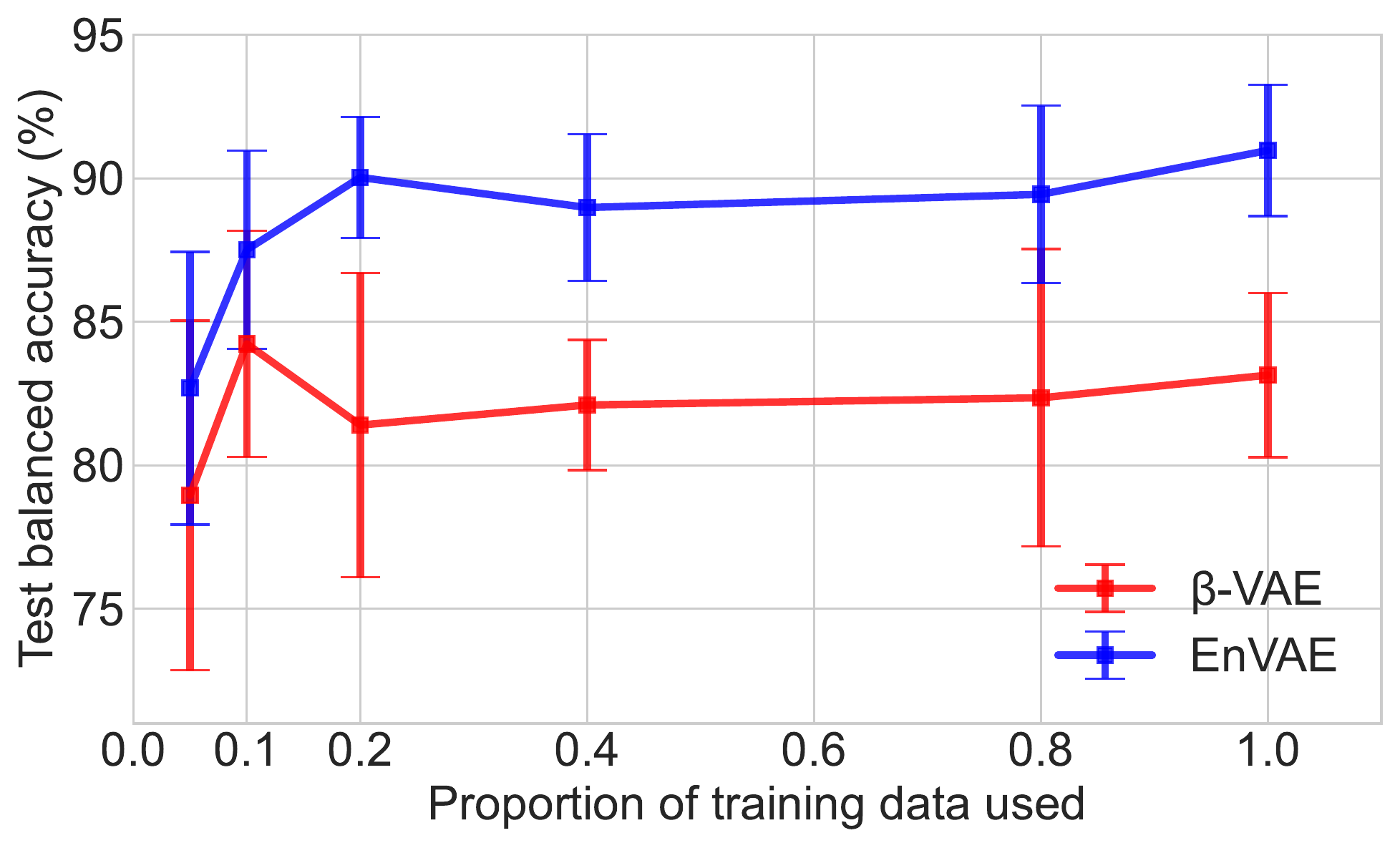}
         \caption{`meta-pam' train size between 60-1200}
     \end{subfigure}
     \hfill
     \begin{subfigure}[b]{0.49\linewidth}
         \centering
         \includegraphics[width=\textwidth]{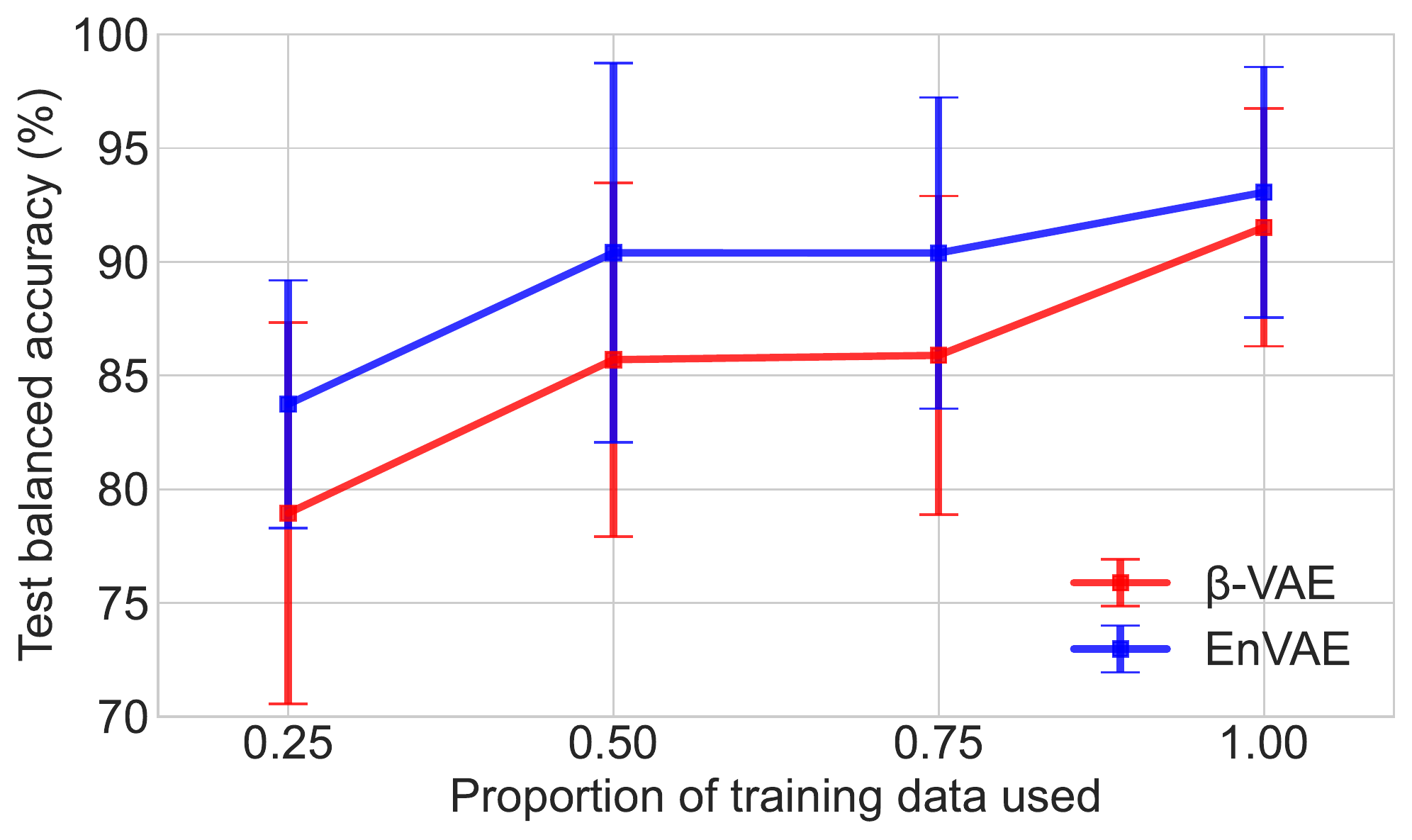}
         \caption{`lung' train size between 30-110}
     \end{subfigure}
     \caption{Benchmarking the learned representations by varying the training set size. We train an MLP classifier on the representations learnt and report the mean~$\pm$~std of the test balanced accuracy. The results indicate that, irrespective of training data volume, EnVAE consistently outperforms a $\beta$-VAE.}
    \label{fig:vary-number-training-samples}
\end{figure}

To improve our understanding, we pose the question, \textit{how does EnVAE perform with differing quantities of training data?} Figure~\ref{fig:vary-number-training-samples} investigates the effect of varying the training set size while maintaining a constant test set. The findings substantiate our earlier observations, demonstrating that EnVAE outperforms current methods, regardless of the training data utilised.

\subsection{Data Exploration using the Learned Representations}
\label{section:umap-plots}

\begin{figure}[thb]
\begin{subfigure}[b]{0.24\textwidth}
         \centering
         \includegraphics[width=\textwidth]{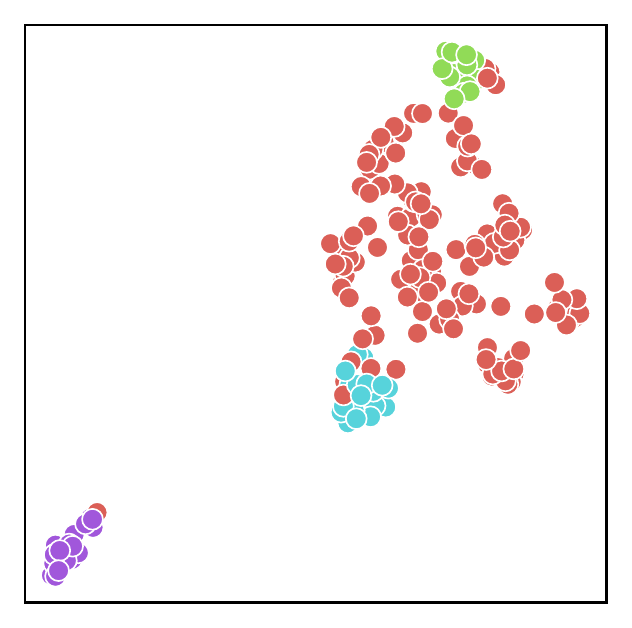}
              \vspace{-16pt}
         \caption{EnVAE (ours)}
     \end{subfigure}
     % \hfill
     \begin{subfigure}[b]{0.24\textwidth}
         \centering
         \includegraphics[width=\textwidth]{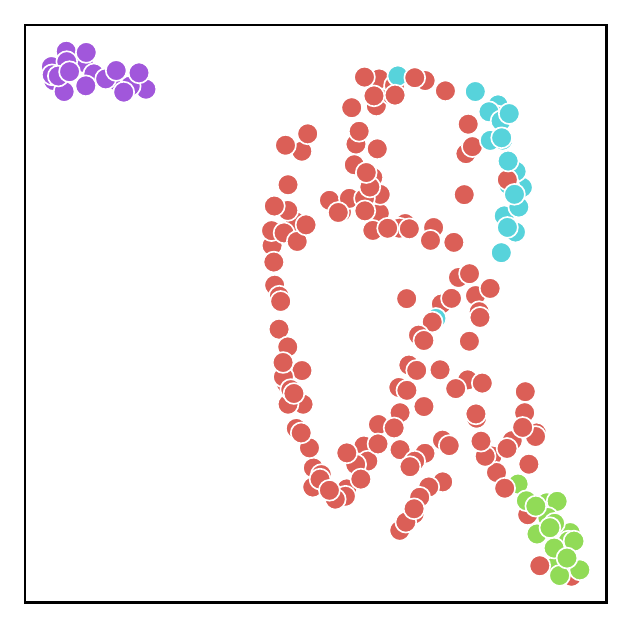}
              \vspace{-16pt}
         \caption{$\beta$-VAE}
     \end{subfigure}
      \begin{subfigure}[b]{0.24\textwidth}
         \centering
         \includegraphics[width=\textwidth]{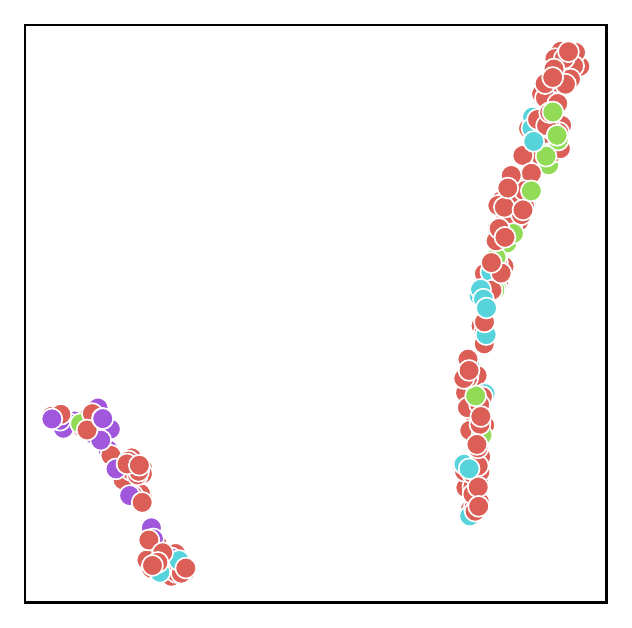}
              \vspace{-16pt}
         \caption{CAE}
     \end{subfigure}
      \begin{subfigure}[b]{0.24\textwidth}
         \centering
         \includegraphics[width=\textwidth]{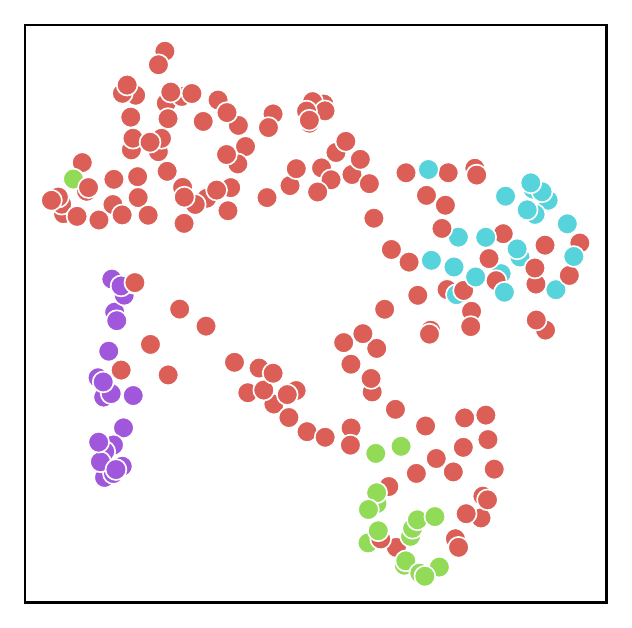}
      \vspace{-16pt}
         \caption{SubTab}
     \end{subfigure}
     % \vspace{-3pt}
     \caption{EnVAE can be used for data exploration. We plot using UMAP the latent representations on `lung', where each point represents the embedding of one sample, and the colour represents its label. The representations learned using EnVAE show that the samples cluster in separated clusters.}
    \label{fig:umap}
\end{figure}

\looseness-1
Our previous findings indicate that EnVAE's representations surpass those produced by current methods. Consequently, we explore whether these representations could provide insights into the data. Utilising UMAP~\cite{mcinnes_umap_2018}, we visualise the latent representations learnt by EnVAE and other benchmark methods. As depicted in Figure~\ref{fig:umap}, the most noticeable clustering is observed within EnVAE's representations, potentially explaining its superior downstream accuracy (as detailed in Section~\ref{section:accuracy-downstream}). CAE's subpar performance could be attributed to its poor convergence in the feature selection mechanism, possibly due to insufficient training samples impeding the identification of relevant features.

% However the tradeoff here is clear, increasing the number of experts causes the resources used by the ELBO to increase exponentially $\mathcal{O}(2^M)$. Furthermore, as the number of features in each group decreases, we can expect a given subset $\pmb{x}_A \in \pmb{x}$ to be less informative of $\pmb{x}$, hence it is not necessarily desirable to to encourage $q_{\phi}(\pmb{z}| \pmb{x}_A)$ to approximate $p(\pmb{z}| \pmb{x})$.\todo{might need to rethink or reword this section} 

\subsection{Dissecting the EnVAE Model}
\label{section:dissecting-model}

\begin{figure}[b!]
    \centering
    \includegraphics[width=\linewidth]{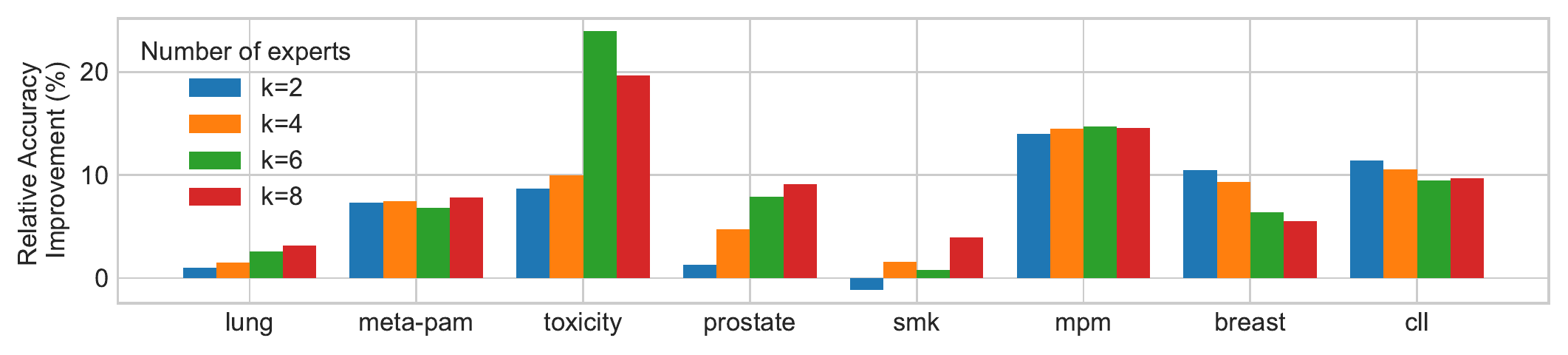}
    \caption{Evaluating the impact of the number of experts on the quality of latent representation learnt in EnVAE. We compare the improvement in balanced accuracy yielded by EnVAE by varying the number of experts $=\{2,4,6,8\}$ over a standard $\beta$-VAE. We observe that for most datasets, 6 or 8 experts yield higher quality latents.}
    \label{fig:improvement}
\end{figure}

\paragraph{Impact of the number of experts.} We first investigate the effect of the number of experts (and thus feature subsets) on the performance of EnVAE in an unsupervised setting. Similar to the classification experiments (Section~\ref{section:accuracy-downstream}) we evaluate via downstream performance of an MLP trained using representations of EnVAE with varying the number of experts $k=\{2,4,6,8\}$. We measure their relative performance (in terms of accuracy) to the one of trained on representations produced by a $\beta$-VAEs with an optimal parameter $\beta$. From the results in Figure~\ref{fig:improvement} we observe that for most datasets, the best performance improvement is achieved by having a larger number of experts (usually 6 or 8). We also note that in all datasets except `smk' (with $k=2$), EnVAE is able to outperform a $\beta$-VAE regardless of the value of $k$. 

We attribute these performance improvements and explain the advantage of EnVAE wrt.\ its relationship with $k$ from the perspective of disentanglement, which has been shown to be important for high-quality representation learning~\cite{bengio_representation_2013,lecun_deep_2015,tschannen_recent_2018}. In this context, a learnt latent representation should contain all the information present in the original data in a distilled, interpretable structure whilst remaining independent from the desired task (e.g., classification). A disentangled representation should separate distinct factors of variation in the data~\cite{bengio_representation_2013}. To assess these properties we estimate the Total Correlation (TC) over the latent representation learnt in both full and (extremely) low sample settings for each dataset. TC measures the dependence among the random variables obtained from the approximate posterior. It equals to $\text{TC} = \mathbb{KL} (  q(\pmb{z}) || \prod_{l=1}^L q(\pmb{z}_l))$, which represents the mutual information between the marginal distributions $q(\pmb{z}_l)$ in each dimension, $l$, of the latent. As computing mutual information is intractable, we approximate it by using fitted Gaussians~\cite{locatello_sober_2020} (see Appendix~\ref{appendix:total-correlation} for complete details). Figure~\ref{fig:tc} shows the results for `lung' and `meta-pam' datasets. Plots for other datasets can be found in Appendix~\ref{app:tc2}. 

Our analysis reveals that in almost every case, regardless of the model used, latent representations learnt in the low-sample setting exhibit higher TC values than in the full-sample setting. This is in line with common consensus that models learn better distinct factors of variation in the data with more samples. Importantly, for most datasets we observe that the TC values generally decrease as the number of experts is increased and eventually plateau, which we advocate provides strong evidence for the benefit of our divide-and-conquer approach. In short, EnVAE improves the disentanglement of latent representations over standard approaches, which we hypothesise in turn explains the superior accuracy achieved in downstream classification tasks.

\begin{figure}[t]
    \centering
     \begin{subfigure}[b]{0.48\textwidth}
         \centering
         \includegraphics[width=\textwidth]{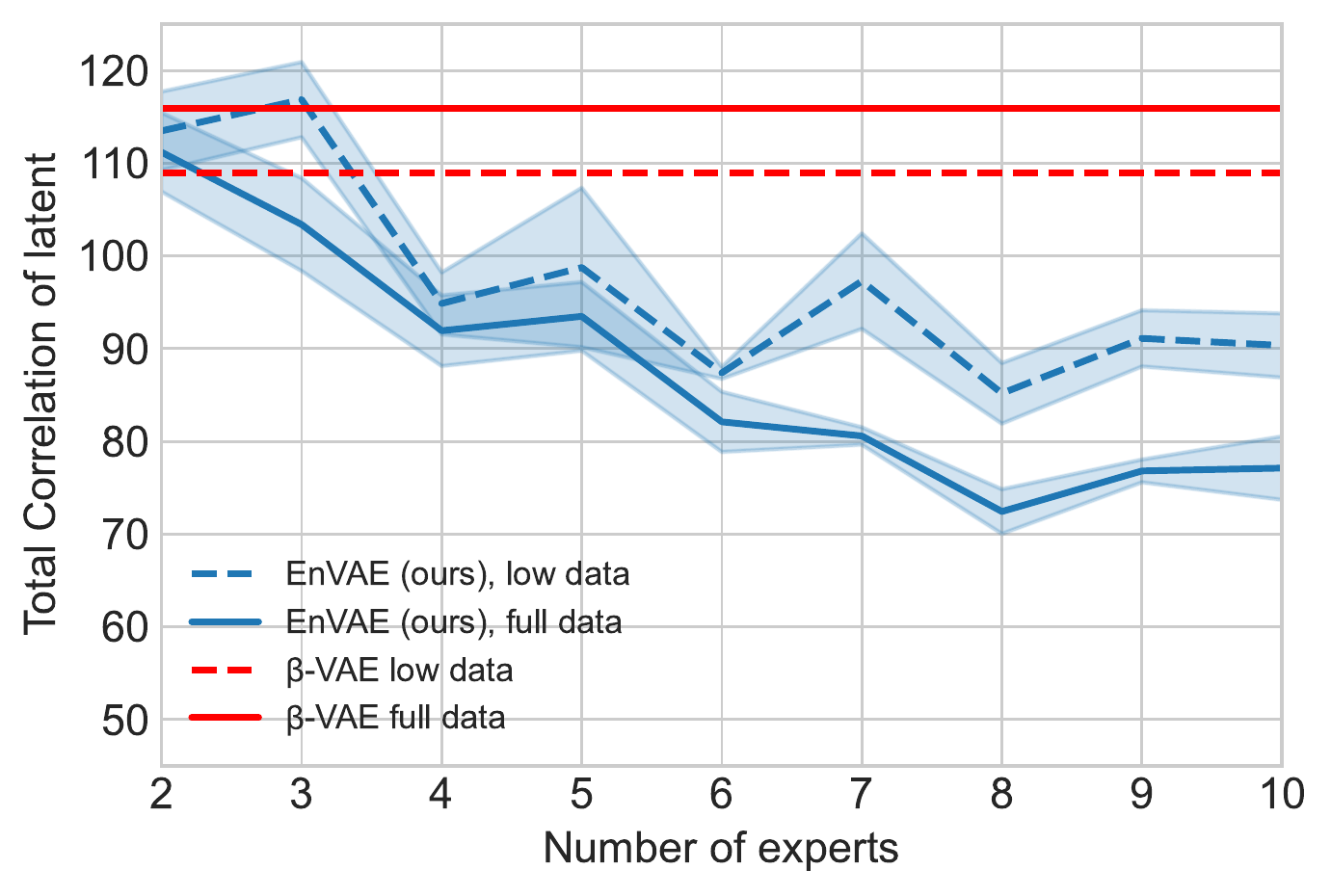}
         \vspace{-18pt}
         \caption{lung}
     \end{subfigure}
          \begin{subfigure}[b]{0.48\textwidth}
         \centering
         \includegraphics[width=\textwidth]{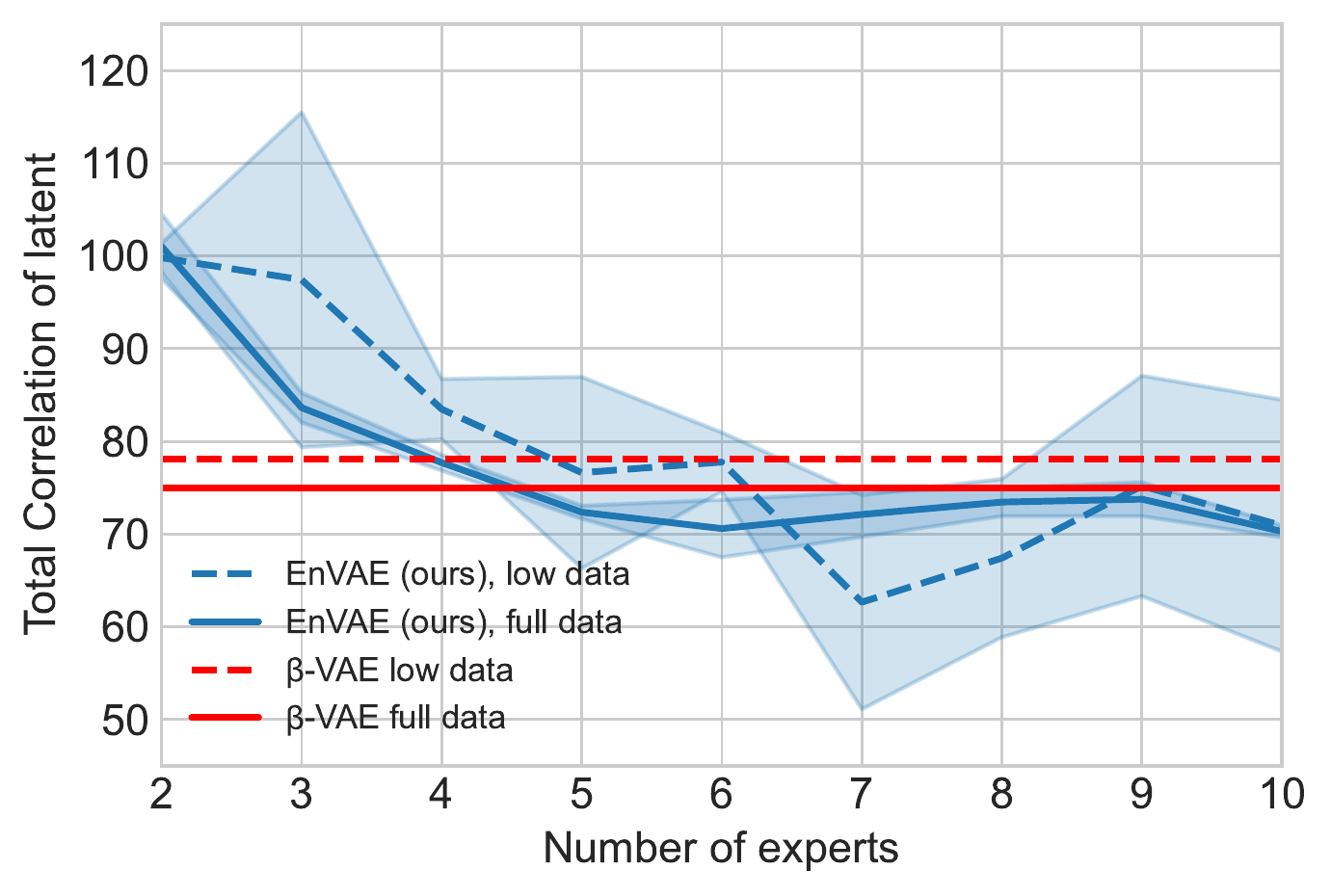}
         \vspace{-18pt}
         \caption{meta-pam}
     \end{subfigure}
     \vspace{-3pt}
    \caption{Comparing the quality of the latent representations learnt by EnVAE and a $\beta$-VAE whilst varying the size of the dataset. We plot the Total Correlation, which measures the disentanglement of the latent representations (lower value in better) for (a) `lung' (b) `meta-pam'. Two observations are notable: (i) Almost all configurations of EnVAE achieve better disentanglement compared to $\beta$-VAE at both levels of training data. (ii) The degree of disentanglement in the representations is enhanced with an increased number of experts within EnVAE.}
    \label{fig:tc} %\todo{add a ,b and datasets}
\end{figure}

\looseness-1
\paragraph{Impact of missing groups of features.} In real-world applications, it is not unusual for observations to be incomplete or contain partial data. A benefit of the MoPoE decomposition of the posterior is the ability to handle missing groups at inference time. We test the EnVAE's ability to generate latent representations from incomplete data by masking groups in the test set. Specifically, we consider EnVAE with 8 experts and generate latent representations by dropping all possible subsets of feature groups, which are then inputted into a pre-trained classifier. The results from Table~\ref{tbl:groupsMasked} demonstrate that EnVAE performs robustly, losing relatively little performance even when most features are missing.

\begin{table*}[h!]
\caption{The impact of missing features on producing a latent representation. We trained EnVAE ($k=8$) fully unsupervised and then generated latent representations of the test set with varying numbers of feature groups missing. We report the balanced accuracy measured by feeding these as input to a pre-trained classifier averaged over all possible combinations of missing groups and averaged over 5 folds. Importantly, we observe that the degradation in latent quality inferred is small even when most features are missing.}

\centering
\resizebox{\textwidth}{!}{
\begin{tabular}{lrrrrrrrr}
    \toprule    
& \multicolumn{8}{c}{Number of groups dropped}  \\
\cmidrule(lr){2-9}
    Dataset & 0 & 1 & 2 & 3 & 4 & 5 & 6 & 7  \\
    \cmidrule(r){1-9}
lung &  $ 94.67 \pm 2.29$   & $ 92.78 \pm 5.33$&  $ 92.21 \pm 5.10$&  $ 91.56 \pm 5.14$&  $ 90.88 \pm 5.02$&  $ 90.29 \pm 5.19$&  $ 89.15 \pm 5.34$&  $ 85.63 \pm 7.27$\\
meta-pam &  $ 90.97 \pm 2.29$ & $ 90.82 \pm 9.49$&  $ 90.79 \pm 9.37 $&  $ 90.53 \pm 9.37$&  $ 90.49 \pm 9.38$&  $ 89.83 \pm 9.38 $&  $ 89.25 \pm 9.62$&  $ 86.81 \pm 10.47$\\
toxicity &  $ 74.75 \pm 8.43$ &  $ 74.75 \pm6.47 $&  $ 74.55 \pm 8.04$&  $ 72.22 \pm 7.92$&  $ 68.01 \pm 8.48$&  $ 64.44 \pm 9.69$5&  $ 57.99 \pm10.58 $&  $ 45.85 \pm 11.13$\\
prostate &  $ 81.27 \pm 6.87$  & $ 82.41 \pm 6.04$&  $ 81.01 \pm 6.02$&  $ 79.04 \pm 8.02$&  $ 80.43 \pm 7.48$&  $ 76.96 \pm 8.54$&  $ 73.12 \pm9.61 $&  $ 70.13 \pm 12.85$\\
smk &  $ 67.15 \pm 8.14$ & $ 66.68 \pm 9.22$&  $ 66.55 \pm 9.05$&  $ 65.51 \pm 8.45$&  $ 64.41 \pm 10.03$&  $ 62.34 \pm 9.19$&  $ 62.11 \pm 9.50$&  $ 58.37 \pm 10.61$\\
mpm &  $ 95.21 \pm 95.21$ &  $ 95.51 \pm4.68 $&  $ 94.82 \pm 4.60$&  $ 94.75 \pm 4.97$1&  $ 93.89 \pm 5.25$&  $ 92.51 \pm5.97 $&  $ 89.70 \pm 7.10$&  $ 84.59 \pm 10.19$\\
cll &  $ 68.06 \pm 12.61$ &  $ 67.45 \pm 11.06$&  $ 66.41 \pm 11.29$&  $ 65.25 \pm 11.27$&  $ 64.20 \pm 12.20$&  $ 62.52 \pm 11.97$&  $ 60.57 \pm 12.65$&  $ 55.55 \pm 13.15$\\
breast &  $ 86.05 \pm 12.87$ & $ 85.20 \pm 6.17$&  $ 85.24 \pm 7.54$&  $ 83.22 \pm 7.84$&  $ 81.20 \pm 12.88$&  $ 76.99 \pm 11.53 $&  $ 72.72 \pm 13.73$&  $ 63.20 \pm 17.21$\\
  \bottomrule \\
\end{tabular}
}
  \label{tbl:groupsMasked}
\end{table*}

\paragraph{Limitations and opportunities.} Whilst our results are promising, we note a few opportunities for improvement. For instance, we did not explore the tradeoff between the reconstruction and regularisation term loss with the hyperparameter $\beta$, which could promote further disentanglement~\citep{burgess_understanding_2018}, and possibly even better downstream performance. While our model demonstrates robust performance under conditions of partial data, it should be noted that this is contingent upon the availability of all features for a given group to be viable. Furthermore, groups were formed by randomly assigning features which may not be optimal. Therefore, we note that there could exist better assignments of features to groups than random selection. In particular, \cite{bouchacourt_multi-level_2018} showed that grouping features which share a common value for a factor of variation resulted in better disentanglement in the latent representation. We attempted to apply similar logic to tabular datasets by using composite feature selection to learn optimal feature groups as proposed in~\cite{imrie_composite_2022}. However, we observed large training instabilities when trying to incorporate such an approach, which ultimately did not lead to any improvements over the presented results. We attribute this to (i) the large number of extra parameters introduced by the feature-selection gate mechanisms and (ii) the in-congruence between the objectives of the group selectors and individual VAEs.

% Thus a few randomly selected missing features are likely to cause all groups to be unavailable. This limitation can be addressed by either increasing the number of groups or forming groups out of features that are semantically co-available for a given inference step by using some expert domain knowledge (e.g., values recorded by the same device). Secondly we note that there is likely to be a better assignment of features to groups than random selection. In particular, Bouchacourt et al.~\cite{bouchacourt_multi-level_2018} who proposed the Multi Level VAE - a latent model that similarly aggregates latents inferred from groups of features - 

\section{Related Work}
\label{section:related-work}
\textbf{Learning high-quality latent representations from HDLSS datasets.} Recent efforts to improve representation learning for tabular data include SubTab~\cite{ucar_subtab_2021} and TabNet~\cite{arik_tabnet_2021}. Similar to our scheme, SubTab divides input features into subsets and auto-encoder attempts to reconstruct the entire set of features from each subset. 
In particular, they hypothesise that reconstructing data from a subset of features can better capture its underlying latent representation which also motivates our approach. However, SubTab differs from our scheme in that a shared encoder and decoder is used for each subset, as well as a different method of aggregation to combine representations.  Whilst parameter sharing is an effective method to reduce the number of parameters and complexity of a model, we suspect that the auto-encoder may struggle to learn the different combinations of the subsets when few samples are available. TabNet~\cite{arik_tabnet_2021} on the other hand uses sequential attention to select a subset of meaningful features at each point of inference. Whilst TabNet grants greater interpretability and performs well on large tabular datasets, we note that it struggles when fewer samples are available~\cite{margeloiu_weight_2022} which is why we did not consider it in our study. 
% \Andrei{@Nav, mention in detail how we are different than SubTab. Also, consider making a table of Related with 3-5 methods, as we do in the WPFS paper. This way, you'll highlight how we are different than (i) multi-modal VAEs, and (ii) SubTab.}

\textbf{Other neural methods for HDLSS scenario.} Many works identify overparameterisation as a key issue that often leads to overfitting and propose various ways of reducing the number of parameters. DietNetworks~\cite{romero_diet_2016} use auxiliary networks to predict the large weight matrices that occur in the first layer of a feed-forward network. FsNets~\cite{singh_fsnet_2020} combine this technique with Concrete Autoencoders~\cite{balin_concrete_2019} by using weight-predictor networks to reduce the number of parameters in the selection and reconstruction layers. WPFS~\cite{margeloiu_weight_2022} takes extends these ideas, using two lightweight  auxiliary networks to compute the first hidden layer's weights and perform global feature selection. However, in this paper we take a different approach to HDLSS tasks by using an ensemble of VAE's and as such this work relates to MoE-Sim-VAE~\cite{kopf_mixture--experts_2021} which uses a single encoder and an ensemble of decoder networks with a Mixture-of-Experts architecture to better learn the various modes of the data. This work also relates to multimodal VAEs~\cite{VAEsCancer,sutter_generalized_2021} which also uses an ensemble of VAEs to integrate multi-modality data (e.g. tabular, text and image data) using encoders and decoders for each modality. Our approach differs however in that we focus on single-modality tabular data and applying the useful properties (Section~\ref{sec:method}) of ensemble-based VAEs to HDLSS settings.

% \textbf{Multimodal VAEs} \Andreiinline{This following paragraph needs much rewording.} \Nikolainline{Remove reference to multimodal; call them something else}
% \Matejainline{Maybe Heterogeneous? Integrative?}
% \citep{wu_multimodal_2018, wu_multimodal_2019, shi_variational_2019, sutter_generalized_2021} have been used to integrate multi-modality data (e.g., images, text and tabular data). The general framework is to use an \textbf{ensemble of VAEs} to learn posterior distributions for each modality and then aggregate these per-modality posteriors into a joint posterior distribution. Considering the potential for missing modalities, if the per-modality latent representations were naively aggregated via concatenation then an encoder would be needed for each of the $2^\text{modalities}$ combinations. The family of \textbf{ensemble-based VAEs} address this issue by decomposing the joint distribution as a product \cite{wu_multimodal_2018}, mixture \cite{shi_variational_2019}, or mixture-of-products \cite{sutter_generalized_2021} of the per-modality posteriors. The second issue of multimodal VAE's is the need to represent modality-specific and shared factors, whilst generating semantically coherent samples across modalities\cite{shi_variational_2019}. 

% Our approach, detailed next, is inspired by multimodal VAEs but targets single-modality tabular data. We leverage the high sample efficiency and the ability to learn using an incomplete feature set granted by using an ensemble of VAEs and, thus we name our approach \textbf{En}semble-\textbf{VAE} (\textbf{EnVAE}).

\section{Conclusion}
In this paper we introduced EnVAE, an ensemble-based VAE with a Mixture-of-Product-of-Experts posterior aggregation as a method to achieve robust dimensionality reduction for High-Dimensional Low-Sample Size (HDLSS) tabular data. 
We presented an alternate factorisation of the MoPoE aggregation that induces implicit data augmentation and greater sample efficiency motivating its application to HDLSS tasks. 
Through a series of experiments on eight real-world biomedical datasets, we demonstrated that EnVAE outperforms other popular dimensionality reduction methods, achieving greater balanced accuracy in downstream classifier tasks. 
Furthermore, we measured lower estimated Total Correlation values for the learnt latent representations for EnVAE in comparison to our baselines, suggesting that this method achieves better latent disentanglement at low sample sizes, which has been identified as an important step towards high-quality representation learning.

%\newpage
% \Andreiinline{We need to cite the published version of the papers, not the Arxiv version (e..g, \citep{wu_multimodal_2018} was published at NeurIPS, but we cite the Arxiv version.).}
% \bibliography{references}
\bibliographystyle{unsrtnat}
\setcitestyle{square}
\bibliography{refs}

\newpage
\appendix
%\resetlinenumber
\renewcommand\thefigure{\thesection.\arabic{figure}}
\setcounter{figure}{0}
\renewcommand\thetable{\thesection.\arabic{table}}
\setcounter{table}{0}
\setcounter{equation}{0}

% \section*{Appendix for submission ``Enhancing Representation Learning on High-Dimensional, Small-Size Tabular Data: A Divide and Conquer Method with Ensembled VAEs''}

\section*{Appendix}

\section{Reproducibility}

\label{app:reproducability}
\subsection{Datasets}
\label{appendix:datasets}

\begin{table}[H]
\centering
\begin{tabular}{lllll}
    \toprule    
Dataset name     & Samples & Features & Classes & Samples per class\\
    \cmidrule(r){1-5}
meta-pam      & 1971&   4160       & 2   &  1642, 329                  \\
lung & 197 & 3312 & 4  &    139, 17, 21, 20     \\
cll & 111 & 11340 & 4 &    11, 49, 51      \\
breast & 104&  22293 & 2  &   62, 42      \\
mpm & 181     &  12533 & 2  &   150, 31       \\
toxicity &   171   & 5748 & 4 &  45, 45, 39, 42         \\
smk &   187   &  19993  & 2&     90, 97    \\
prostate & 102 & 5966 & 2 &   50, 52           \\\bottomrule
\end{tabular}
\caption{Details of eight real-world biomedical datasets used in this paper.}
  \label{tbl:app_datasets}
\end{table}
Table~\ref{tbl:app_datasets} summarises all the datasets. Seven of the datasets are open-source and are available online: \textbf{`breast'}~\cite{chowdary_prognostic_2006}, \textbf{`mpm'}~\cite{gordon_translation_2002} can be found at \hyperlink{https://github.com/ramhiser/datamicroarray}{https://github.com/ramhiser/datamicroarray}. : \textbf{CLL-SUB-111} (called \textbf{‘cll’})~\cite{haslinger_microarray_2004}, \textbf{lung}~\cite{bhattacharjee_classification_2001}, \textbf{Prostate GE} (called \textbf{‘prostate’})~\cite{singh_gene_2002}, \textbf{SMK-CAN-187} (called \textbf{‘smk’})~\cite{spira_airway_2007}
and \textbf{TOX-171} (called \textbf{‘toxicity’})~\cite{bajwa_cutting_2016} can be found at \hyperlink{(https://jundongl.github.io/scikit-feature/datasets.html)}{(https://jundongl.github.io/scikit-feature/datasets.html)}. \textbf{meta-pam} is derived from the \textbf{METABRIC} dataset~\cite{metabric_group_genomic_2012} by combining the molecular data with the clinical label ‘Pam50Subtype’. Because the label ‘Pam50Subtype’ was very imbalanced, the task was transformed into
a binary task of basal vs non-basal by combining the classes ‘LumA’, ‘LumB’, ‘Her2’, ‘Normal’ into one class and using the remaining class ‘Basal’ as the second class. The Hallmark gene set associated with breast cancer was selected and the final dataset contained 1971 samples forming the ``largest" of the small-sample size datasets we looked at, allowing us to more granularly investigate the effect of decreasing the number of training samples.

\textbf{Data preprocessing} Before training the models, we normalise the dataset by scaling all features to be between 0 and 1 using \texttt{scikit-learn.preprocessing.MinMaxScaler}.

\subsection{Computing Resources}
We trained over 20000 models on a Tesla P100-PCIE with 16GB memory with a Intel(R) Xeon(R) Gold 6142 CPU @ 2.60GHz 16-core processor. The operating system was Ubuntu 20.04.4 LTS. A small portion of the training (<5\%) was also performed using a RTX 3090 with 24GB memory with a Xeon® E5-2609 v3 @ 1.90 GHz 16-core processor. We estimate that $\sim 400$ GPU-hours were used to run all experiments.

\subsection{Training details, Hyperparameters}
\label{app:hyperparameters}
\textbf{Software Implementation} We implemented all models using PyTorch\cite{paszke_pytorch_2019}, an open-source deep learning library with a BSD licence. All numerical plots and graphics have been generated using Matplotlib, a Python-based plotting library with a BSD licence. The model architecture Figure~\ref{fig:model_architecture} was generated using 
\href{https://github.com/jgraph/drawio}{draw.io}, a free drawing software under Apache License 2.0. Our submitted code contains the exact version of all software and libraries we use.

We attach our code to this submission, and we will release it under the MIT licence upon publication.

\textbf{Training details} Below we present the most important training settings for each model. 
\begin{itemize}
    \item $\beta$-VAE and CAE  each use encoder and decoder networks with two layers respectively of size 128,128. 
    \item For $\beta$-VAE we use a monotonic $\beta$ schedule, starting with $\beta=0$ at the start of training and linearly increasing up to its maximum value after 100 epochs.
    \item For CAE, we use an exponential annealing schedule with a duration of 300 epochs from 10 to 0.01, as suggested in ~\cite{balin_concrete_2019}.
    \item Each model was trained for 10000 iterations with early stopping on the validation reconstruction loss / cross-entropy and a patience of 150 epochs. 
    \item The classifier is an MLP with two hidden layers, each of size 64. We used the same architecture for evaluating downstream performance and when jointly trained with each model.
\end{itemize}
\textbf{Hyperparameters.} We kept batch size and learning rate constant across all models at 8 and 0.001 respectively. To prevent gradient underflow / overflow we also used gradient clipping at 2.5.  We kept the architecture of the classifier used for evaluation constant throughout with two hidden layers of size 64,64, ReLU activations, batch normalisation layers and dropout(p=0.5). For the $\beta$-VAE baseline we grid searched $\beta \in \{0.125,0.25,0.5,1,2,4,8,16\}$ and performed five-fold cross-validation, selecting optimal values based on validation loss (Fig. \ref{tbl:beta}). For the SubTab baseline we mainly use the suggested values from the original paper~\citep{ucar_subtab_2021}. In particular we use 4 subsets with an overlap of 75\%,  swap noise with a masking ratio of 0.3. We found that enabling contrastive loss led to training instabilities in for all datasets and thus only enabled distance loss. For CAE, we grid-searched the number of selected features $\in \{100,200,300\}$ and found 300 to give the best and most stable results in all datasets.

\begin{table}[H]
\centering
\begin{tabular}{lcccccccc}
    \toprule    
    Dataset & meta-pam &lung & mpm & smk & cll & breast & toxicity & prostate  \\
    % \cmidrule(r){1-5}
    \midrule
$\beta$     & 16     & 0.125 &  0.25& 1 &0.125 & 0.25& 0.125 &0.125             \\\bottomrule
\end{tabular}
\caption{Optimal values of $\beta$ for $\beta$-VAE found via five-fold cross-validation for each dataset}
  \label{tbl:beta}
\end{table}

\section{Computing Total Correlation}
\label{appendix:total-correlation}

We choose to measure disentanglement based on Total Correlation (TC)\cite{watanabe_information_1960} which has solid theoretical grounding in Information Theory by measuring the mutual information between the marginal distributions each dimension of the $\pmb{z}$ (Eq. \ref{eq:tc})
\begin{equation}
    TC = \mathbb{KL} \left(  p(\pmb{z}) || \prod_{j=1}^D p(\pmb{z}_j) \right)
    \label{eq:tc}
\end{equation}
However as TC is intractable to compute, it has to be estimated, which we do by using a fitted Gaussian as it provides a robust method to detect if a representation is not factorising based on its first and second moments \cite{locatello_sober_2020}. In particular it avoids common issues like underestimation of the true value and dependence on training a separate discriminator such as $\beta$-VAE's metric and FactorVAE's metric\cite{kim_disentangling_2018}. It also inherently works well in a probabilistic setting as opposed to MIG\cite{chen_isolating_2018} which does not make use of the encoder's distribution. 

\begin{equation}
    TC = \mathbb{KL} \left(  \mathcal{N}(\mu_{\pmb{z}}, \Sigma_{\pmb{z}}) || \prod_{j=1}^D \mathcal{N}( \mu_{\pmb{z}_j}, \Sigma_{\pmb{z}_j})  \right)
    \label{eq:tc_gaussian}
\end{equation}

To compute the TC based on a fitted Gaussian, first we obtain the latent representations for each sample in the dataset, $\pmb{z}$ by either sampling the posterior distribution or taking the mean. We then compute the mean latent representation $\mu_{\pmb{z}}$ and the covariance matrix $\Sigma_{\pmb{z}}$ and then compute the total correlation of a multivariate Gaussian with mean $\mu_{\pmb{z}}$ and covariance matrix $\Sigma_{\pmb{z}}$ as Eq. \ref{eq:tc_gaussian}. Here, $j$ indexes the dimensions in the latent space.

\section{Classification Performance Varying Dataset Size}
\label{appendix:classification_performance}

We assess the downstream classification accuracy of an MLP classifier trained on the extracted latent representations. After training the models to learn representations for each dataset unsupervised, we compute the latent representations for the entire dataset. Then, we train an MLP classifier five times with different seeds for each VAE model -- while maintaining the same dataset splits -- and resulting in 25 total runs. Additionally, we measure the performance of EnVAE with varying numbers of experts$=\{2,4,6,8\}$. In Tables \ref{tbl:pam50_summary}-\ref{tbl:cll_summary} we report the mean~$\pm$~std of the test balanced accuracy averaged across all 25 runs in both full and low training samples settings.

\begin{table}[H]
\centering
\begin{tabular}{lcccl}
    \toprule    
& \multicolumn{2}{c}{Full data(n=1261)}  &\multicolumn{2}{c}{Low data(n=58)}  \\
\cmidrule(lr){2-3}\cmidrule(lr){4-5}
    Model & mean& std & mean& std  \\
    \cmidrule(r){1-5}
$\beta$-VAE      & 83.14     & 2.86  &  77.49 & 3.70             \\
CAE      &73.54    &   2.14   & 70.25 &  4.49       \\
SubTab      &  88.11  &   1.51   & 61.45  &   2.24   \\
EnVAE(k=2)      & 90.45     &  2.14   & 79.23  & 4.06          \\
EnVAE(k=4)      & 90.59    &   3.03   & 81.55  & 3.89        \\
EnVAE(k=6)      &  89.93   &  2.09 & 80.47    & 3.54           \\
EnVAE(k=8)      & \textbf{90.97}    & 2.29   & \textbf{81.85}  & 2.89              \\\bottomrule
\end{tabular}
\caption{Summary of performance of models on `meta-pam'.}
  \label{tbl:pam50_summary}
\end{table}

\begin{table}[H]
\centering
\begin{tabular}{lcccl}
    \toprule    
& \multicolumn{2}{c}{Full data(n=118)}  &\multicolumn{2}{c}{Low data(n=29)}  \\
\cmidrule(lr){2-3}\cmidrule(lr){4-5}
    Model & mean& std & mean& std  \\
    \cmidrule(r){1-5}
$\beta$-VAE      & 91.52     & 5.23    & 78.95 & 8.38                  \\
CAE      &  74.31   &  6.48   & 62.33& 16.39                  \\
SubTab      &  89.12  &  7.29    &  78.05 &  10.20    \\
% DietNetwork      &   71.69  &   16.52    &   43.96  &    13.62       \\
EnVAE(k=2)      & 92.56     &   5.09   & 82.88  & 7.34          \\
EnVAE(k=4)      &  94.14     &   5.35   & 83.93  & 5.88        \\
EnVAE(k=6)      &  93.06    &  5.51   & 83.74    & 5.45             \\
EnVAE(k=8)      & \textbf{94.67}     &   6.67   & \textbf{85.13}  & 6.16             \\\bottomrule
\end{tabular}
\caption{Summary of performance of models on `lung'}
  \label{tbl:lung_summary}
\end{table}

\begin{table}[H]
\centering
\begin{tabular}{lcccl}
    \toprule    
& \multicolumn{2}{c}{Full data(n=102)}  &\multicolumn{2}{c}{Low data(n=51)}  \\
\cmidrule(lr){2-3}\cmidrule(lr){4-5}
    Model & mean& std & mean& std  \\
    \cmidrule(r){1-5}
$\beta$-VAE      &  50.83   &    9.31   &  42.44    &  7.51 \\
CAE      &  52.35 &   10.85  &   48.21    &     10.03\\
SubTab      &  74.16  &   8.23   &   59.99&  14.12    \\
EnVAE(k=2)      &  59.51   &    7.42   &  55.32 &   7.75              \\
EnVAE(k=4)      &   60.79  &   7.98    &  57.04      &  10.26      \\
EnVAE(k=6)    & \textbf{ 74.75}   &  8.43     &  66.38 &   6.50            \\
EnVAE(k=8)     &  70.48   &  12.87     & \textbf{ 67.06} &   10.09               \\\bottomrule
\end{tabular}
\caption{Summary of performance of models on `toxicity'}
  \label{tbl:toxicity_summary}
\end{table}

\begin{table}[H]
\centering
\begin{tabular}{lcccl}
    \toprule    
& \multicolumn{2}{c}{Full data(n=61)}  &\multicolumn{2}{c}{Low data(n=15)}  \\
\cmidrule(lr){2-3}\cmidrule(lr){4-5}
    Model & mean& std & mean& std  \\
    \cmidrule(r){1-5}
$\beta$-VAE       &  73.16   & 9.86  &  71.00  &  12.42       \\
CAE      & 56.49  & 8.27     &    59.53    &  9.89\\
SubTab      &  66.34 &   14.47  & 59.25  &  10.22    \\
EnVAE(k=2)     &  74.44  & 11.88  &  75.26 &   8.62                   \\
EnVAE(k=4)     &  77.91  &  9.22 &  \textbf{75.52} &   15.47                  \\
EnVAE(k=6)  &  81.05  &  5.36 &  74.87 &   12.38                   \\
EnVAE(k=8)     &  \textbf{81.27}  &  6.87 & 73.54  &  12.45                \\\bottomrule
\end{tabular}
\caption{Summary of performance of models on `prostate'}
  \label{tbl:prostate_summary}
\end{table}

\begin{table}[H]
\centering
\begin{tabular}{lcccl}
    \toprule    
& \multicolumn{2}{c}{Full data(n=112)}  &\multicolumn{2}{c}{Low data(n=27)}  \\
\cmidrule(lr){2-3}\cmidrule(lr){4-5}
    Model & mean& std & mean& std  \\
    \cmidrule(r){1-5}
$\beta$-VAE      &   63.20  &    10.23   &   58.18   &  10.24 \\
CAE     &   54.37  &   7.82    &   51.41  &    7.33  \\
SubTab      &   62.14 &  5.68    & \textbf{61.07}  &  7.97    \\
EnVAE(k=2)   &  62.02  &   5.11  &  60.15     &   8.28                \\
EnVAE(k=4)       &  64.8   &   8.18    &    57.71   &    12.19            \\
EnVAE(k=6)    &   63.99  &   10.44    & 59.13& 9.32          \\
EnVAE(k=8)      & \textbf{67.15}   &    8.14   &     57.69  &       10.84         \\\bottomrule
\end{tabular}
\caption{Summary of performance of models on `smk'}
  \label{tbl:smk_summary}
\end{table}

\begin{table}[H]
\centering
\begin{tabular}{lcccl}
    \toprule    
& \multicolumn{2}{c}{Full data(n=108)}  &\multicolumn{2}{c}{Low data(n=27)}  \\
\cmidrule(lr){2-3}\cmidrule(lr){4-5}
    Model & mean& std & mean& std  \\
    \cmidrule(r){1-5}
$\beta$-VAE      &   80.53  &  9.96     &77.99       &  14.03\\
CAE      &  79.33   &  11.69     &  60.56     &  17.25\\
SubTab      &  89.08   &  8.83    & 80.99  & 15.91     \\
EnVAE(k=2)   &  94.55   &   5.19    &   89.12    & 9.71 \\
EnVAE(k=4)   &  95.02   &  6.41    & 88.84       &  10.76\\
EnVAE(k=6)    &   \textbf{95.21}  & 4.38  &  \textbf{90.76}  &   7.69  \\
EnVAE(k=8)    &   95.07  &  5.26     &   90.42    & 9.09\\      \bottomrule
\end{tabular}
\caption{Summary of performance of models on `mpm'}
  \label{tbl:mpm_summary}
\end{table}

\begin{table}[H]
\centering
\begin{tabular}{lcccl}
    \toprule    
& \multicolumn{2}{c}{Full data(n=63)}  &\multicolumn{2}{c}{Low data(n=31)}  \\
\cmidrule(lr){2-3}\cmidrule(lr){4-5}
    Model & mean& std & mean& std  \\
    \cmidrule(r){1-5}
$\beta$-VAE      &   75.59  &   10.32   & 72.16      & 18.64 \\
CAE      &   78.00 &   7.08   & 63.96      & 11.81 \\
SubTab      &  65.47  &   16.15   &  66.69 &  17.10    \\
EnVAE(k=2)   &   \textbf{86.05} &   12.87    &  78.53    &  14.83 \\
EnVAE(k=4)   &   84.92  &    12.24   &  \textbf{83.41}     & 9.80\\
EnVAE(k=6)    &   81.97  &   12.88    &    81.95   & 11.68\\
EnVAE(k=8)    &  81.13   &  12.21     &    80.19   & 13.6\\      \bottomrule
\end{tabular}
\caption{Summary of performance of models on `breast'}
  \label{tbl:breast_summary}
\end{table}

\begin{table}[H]
\centering
\begin{tabular}{lcccl}
    \toprule    
& \multicolumn{2}{c}{Full data(n=66)}  &\multicolumn{2}{c}{Low data(n=33)}  \\
\cmidrule(lr){2-3}\cmidrule(lr){4-5}
    Model & mean& std & mean& std  \\
    \cmidrule(r){1-5}
$\beta$-VAE      &  56.64   &  14.07     &  53.75    & 10.35\\
CAE      &  54.35   &  9.10     &  52.65     & 11.43    \\
SubTab      &    54.32&  11.16    & 58.04  &  12.88    \\
EnVAE(k=2)   &  \textbf{68.06}   &   12.61    &   65.69    &10.14 \\
EnVAE(k=4)   &   67.22  &  11.55     &   \textbf{67.64}   &10.49 \\
EnVAE(k=6)    &   66.10  &  11.13     &    65.94   &10.60 \\
EnVAE(k=8)    &   66.35  &   11.73    &     65.25  & 10.30\\      \bottomrule
\end{tabular}
\caption{Summary of performance of models on cll}
  \label{tbl:cll_summary}
\end{table}

\section{Evaluating the Total Correlation Across Models and Dataset Sizes}\label{app:tc2}
\begin{figure}[H]
    \centering
\begin{subfigure}[b]{0.49\textwidth}
    \includegraphics[width=\textwidth]{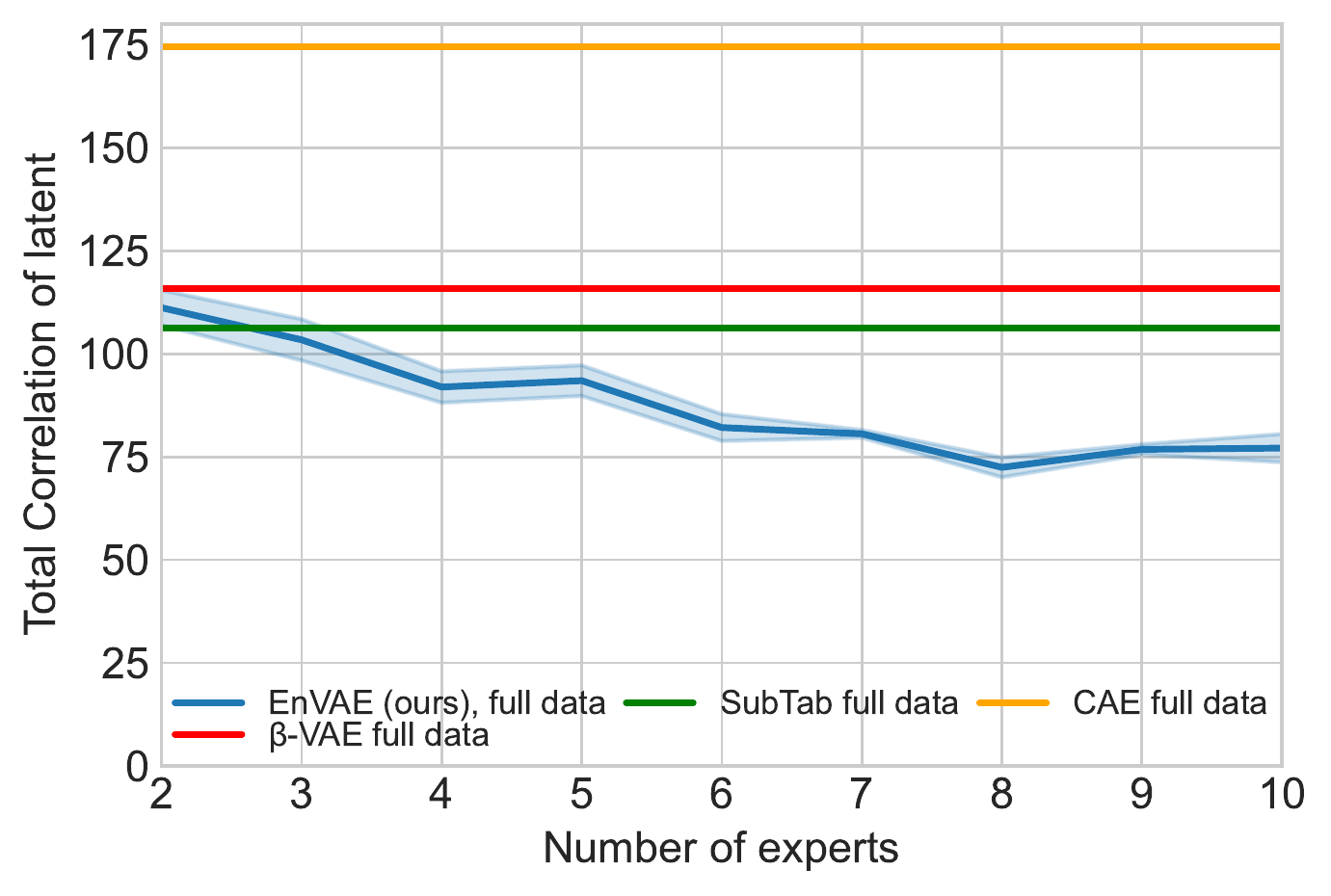}
    \vspace{-15pt}
    \caption{lung}
\end{subfigure}
\begin{subfigure}[b]{0.49\textwidth}
    \includegraphics[width=\linewidth]{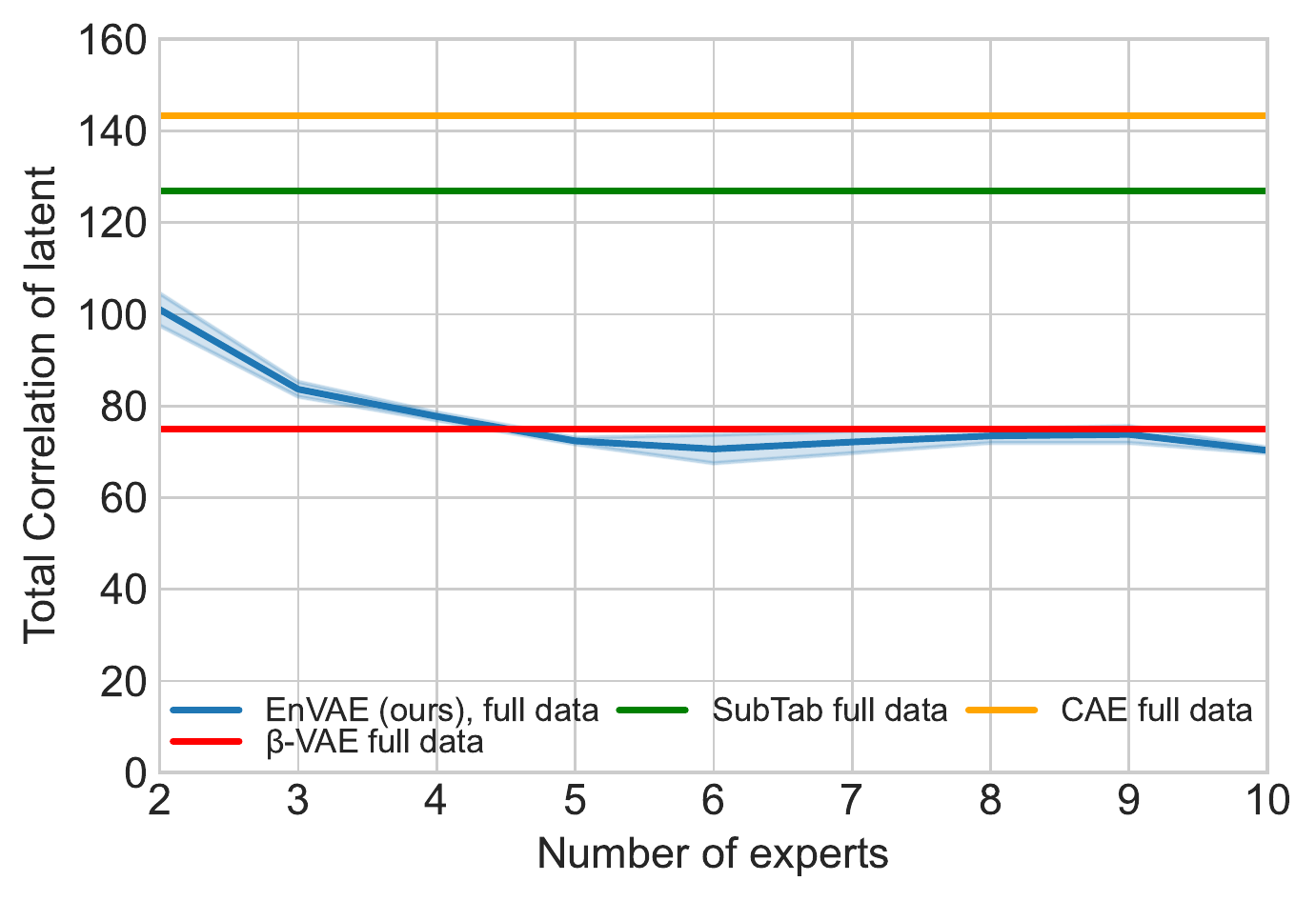}
        \vspace{-15pt}
    \caption{meta-pam}
\end{subfigure}
\begin{subfigure}[b]{0.49\textwidth}
    \includegraphics[width=\linewidth]{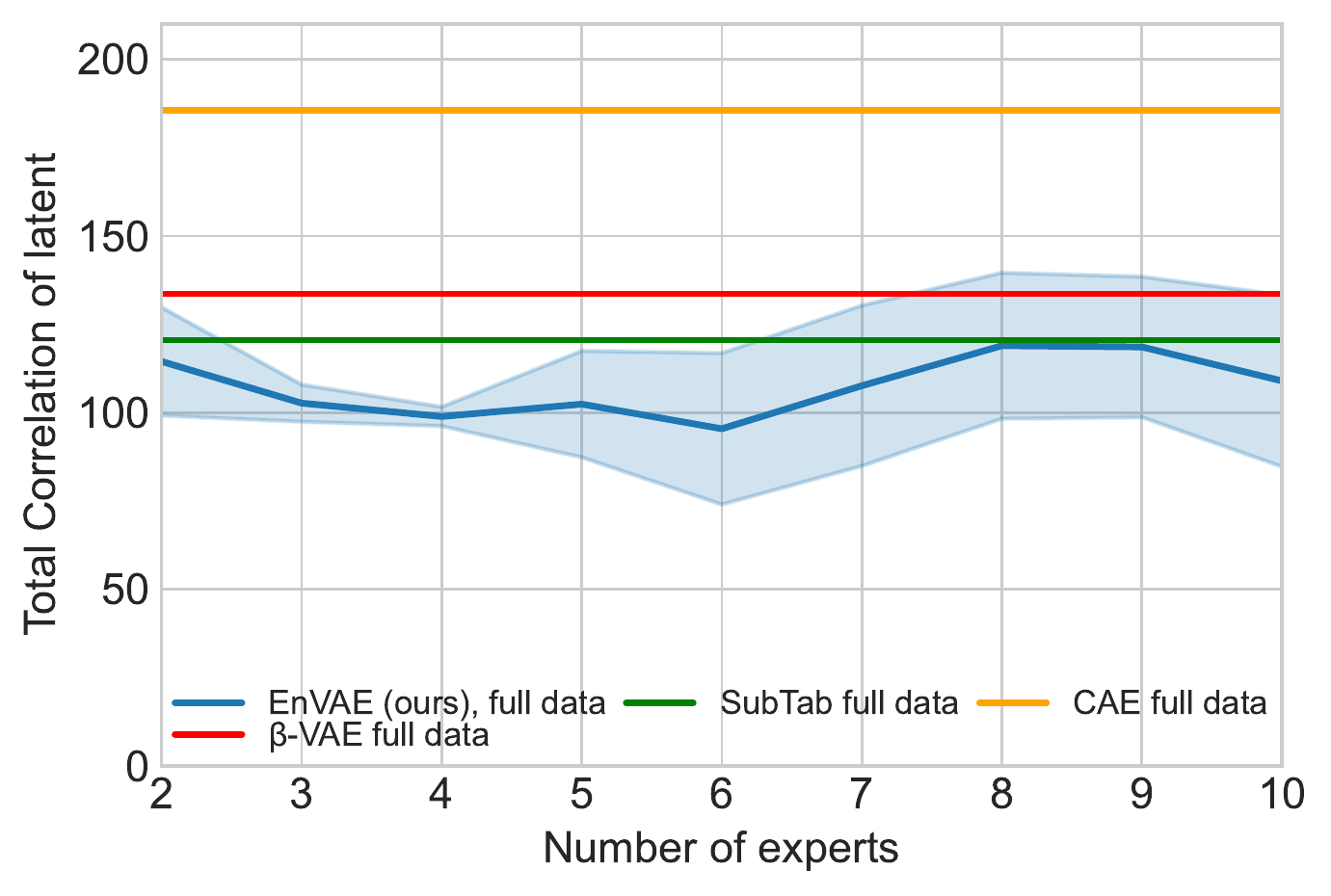}
        \vspace{-15pt}
    \caption{smk}
\end{subfigure}
\begin{subfigure}[b]{0.49\textwidth}
    \includegraphics[width=\linewidth]{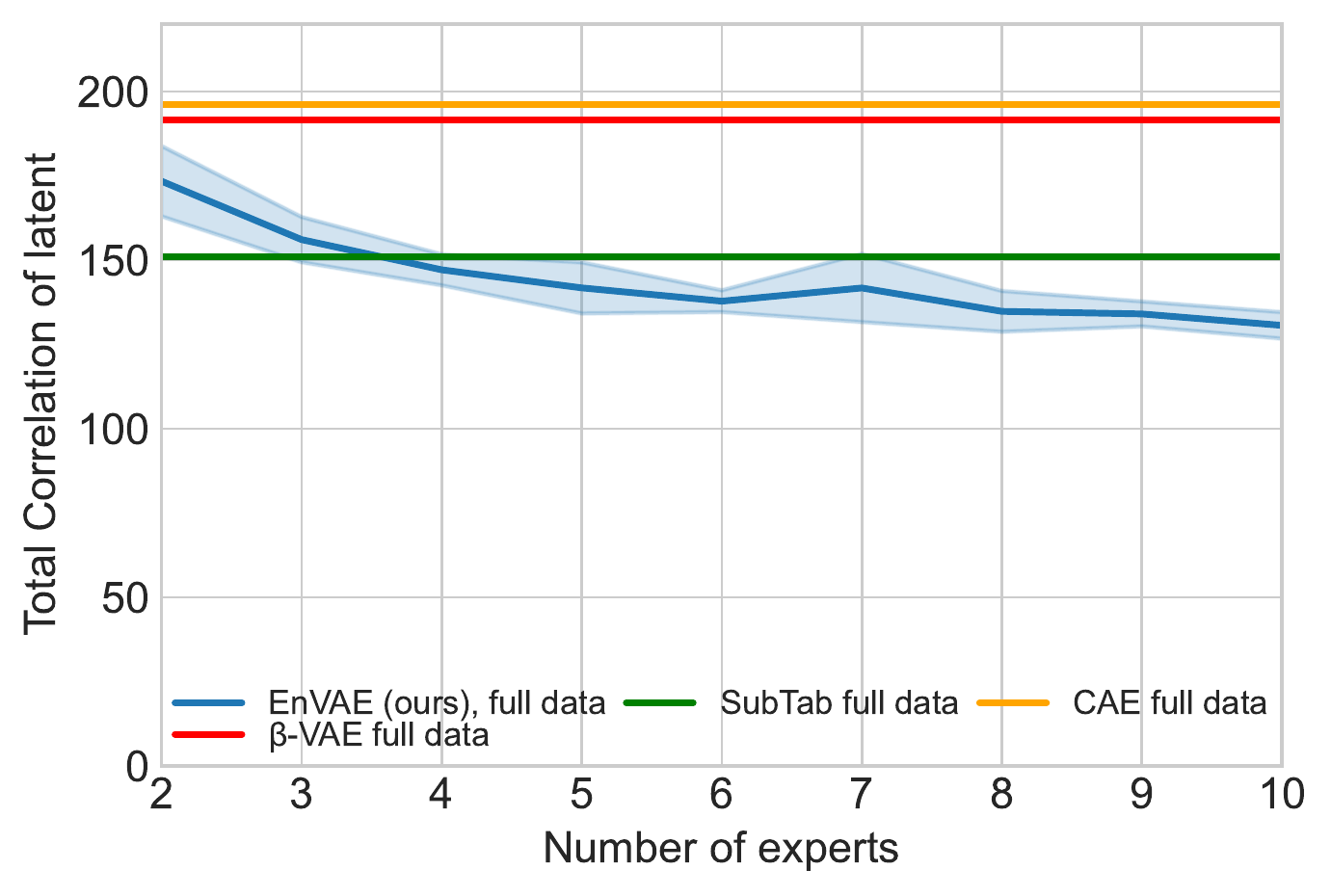}
        \vspace{-15pt}
    \caption{prostate}
\end{subfigure}
\begin{subfigure}[b]{0.49\textwidth}
    \includegraphics[width=\linewidth]{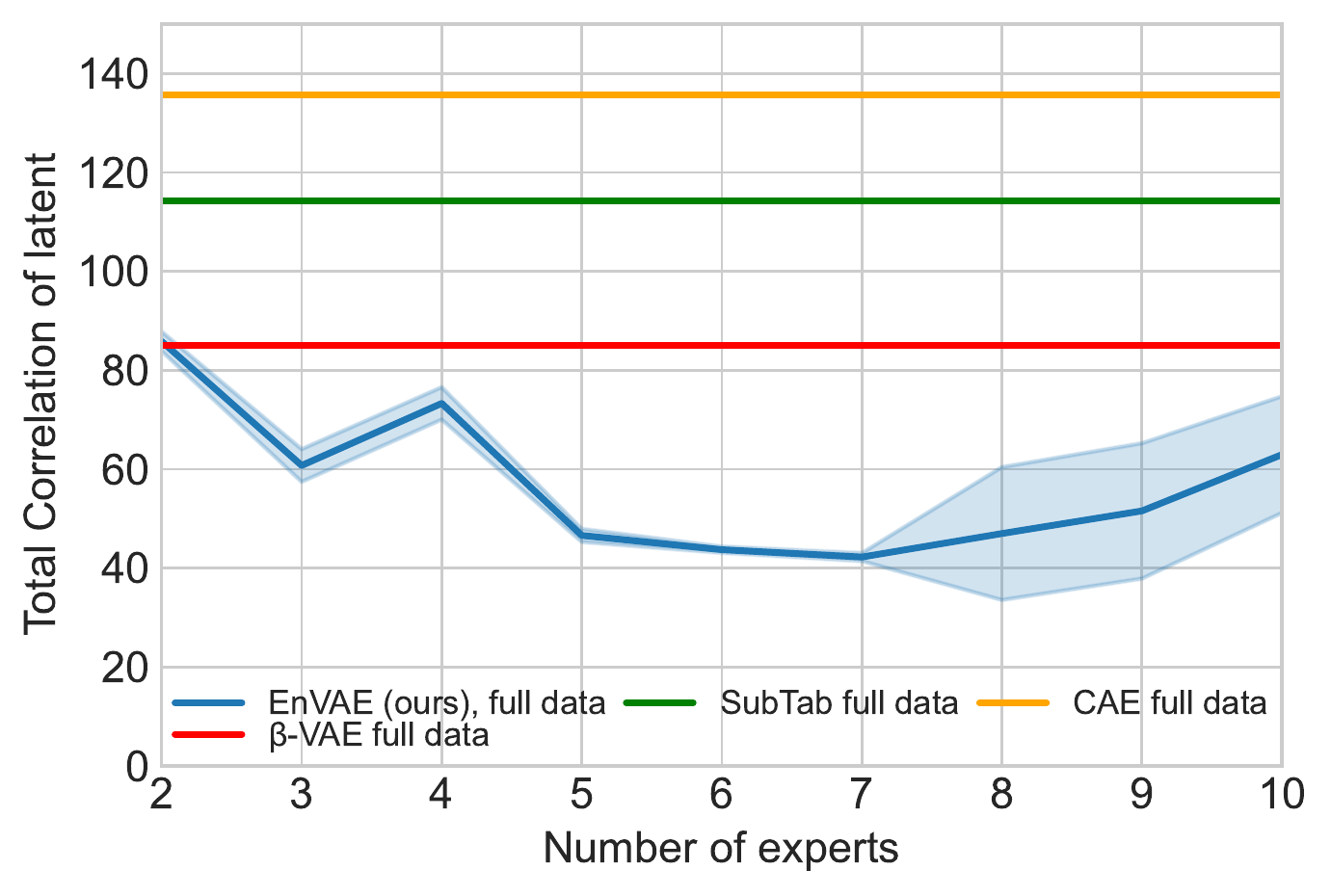}
        \vspace{-15pt}
    \caption{toxicity}
\end{subfigure}
\begin{subfigure}[b]{0.49\textwidth}
    \includegraphics[width=\linewidth]{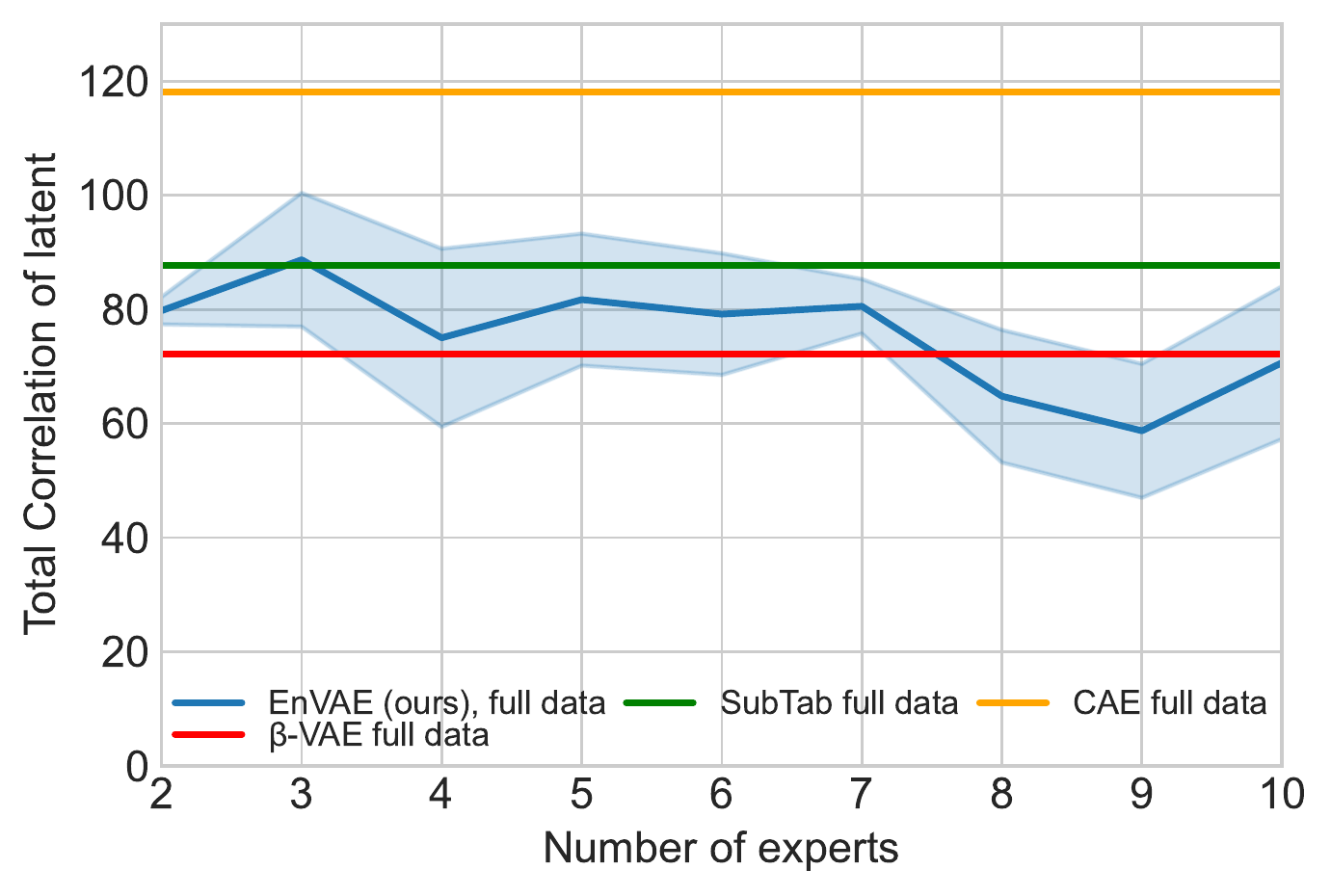}
        \vspace{-15pt}
    \caption{mpm}
\end{subfigure}
\begin{subfigure}[b]{0.49\textwidth}
    \includegraphics[width=\linewidth]{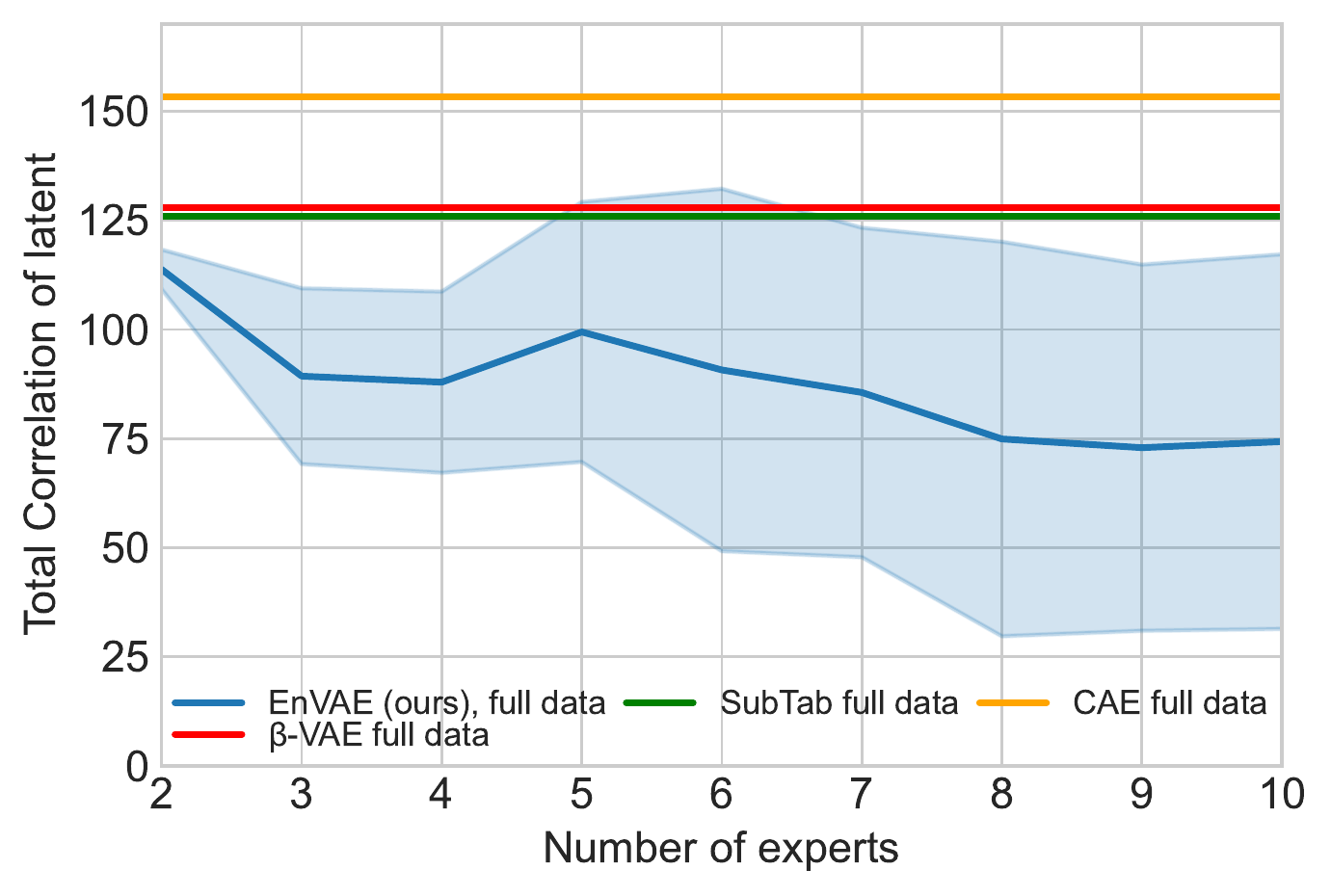}
        \vspace{-15pt}
    \caption{breast}
\end{subfigure}
\begin{subfigure}[b]{0.49\textwidth}
    \includegraphics[width=\linewidth]{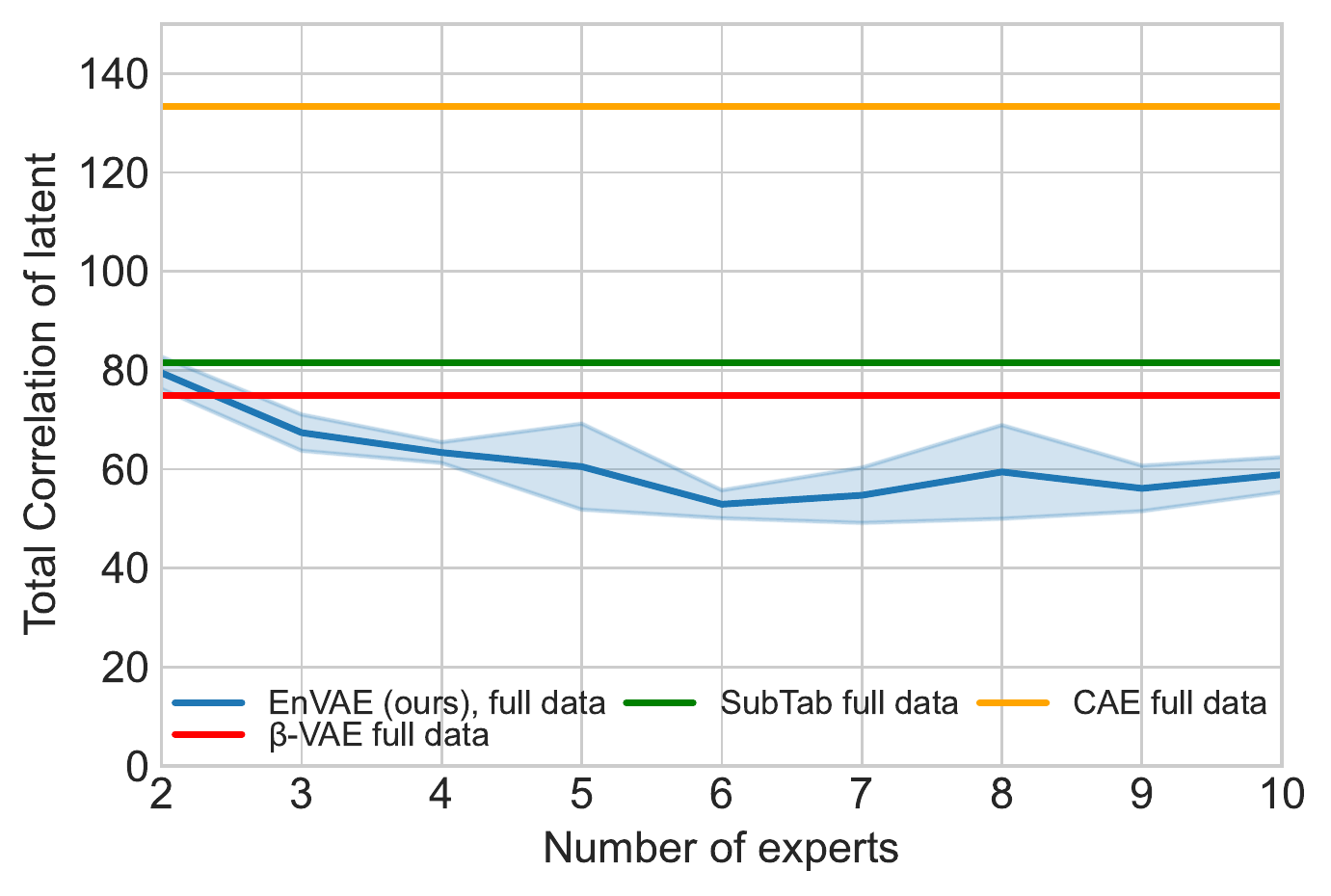}
        \vspace{-15pt}
    \caption{cll}
\end{subfigure}
    \caption{Total Correlation (TC) of the latent representations learnt on the \textit{full datasets} by the benchmark models and EnVAE with varying numbers of experts. EnVAE consistently produces better latent representations, as indicated by a lower TC.}
    \label{fig:app_tc}
\end{figure}

\begin{figure}[H]
    \centering
\begin{subfigure}[b]{0.49\textwidth}
    \includegraphics[width=\textwidth]{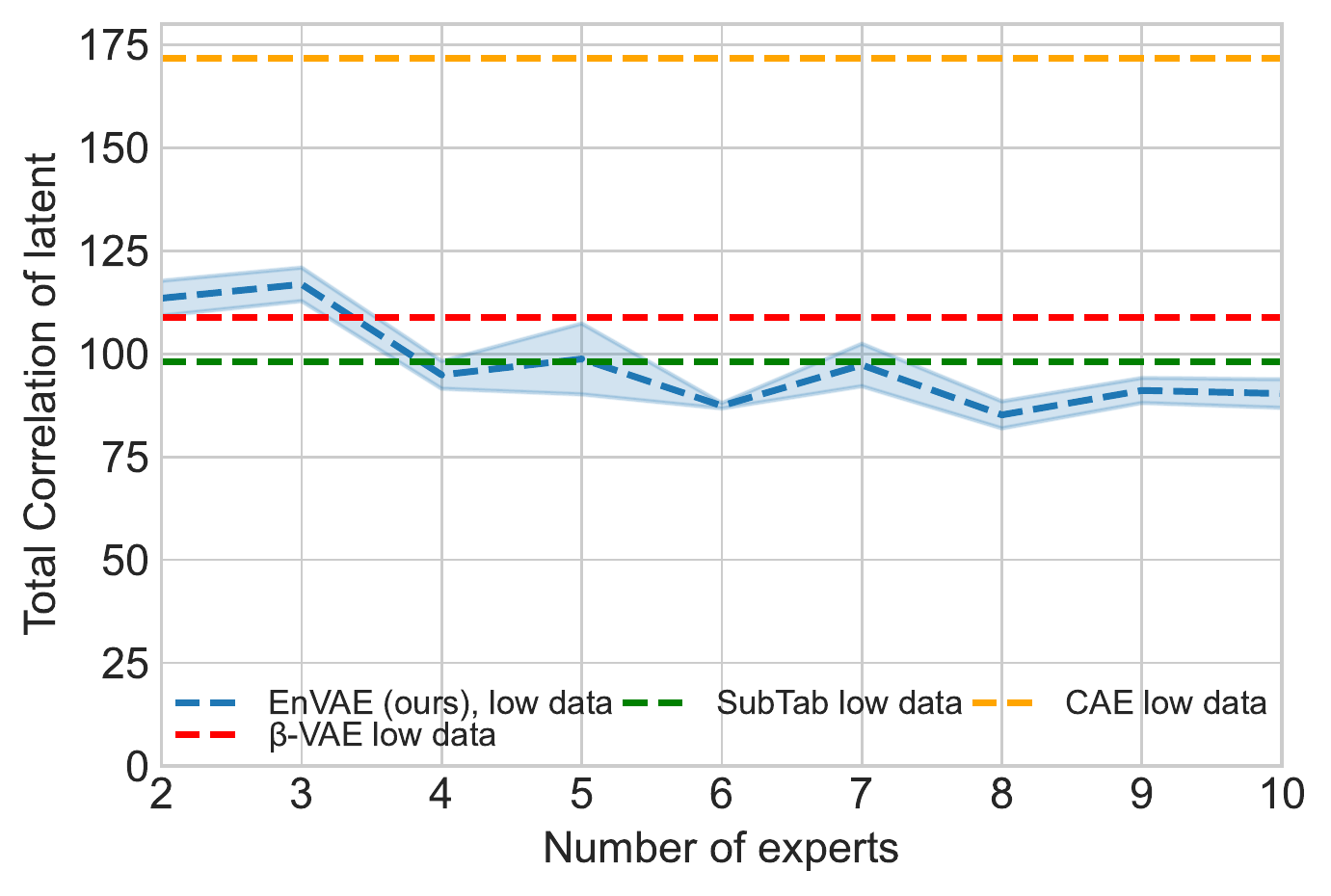}
    \vspace{-15pt}
    \caption{lung}
\end{subfigure}
\begin{subfigure}[b]{0.49\textwidth}
    \includegraphics[width=\linewidth]{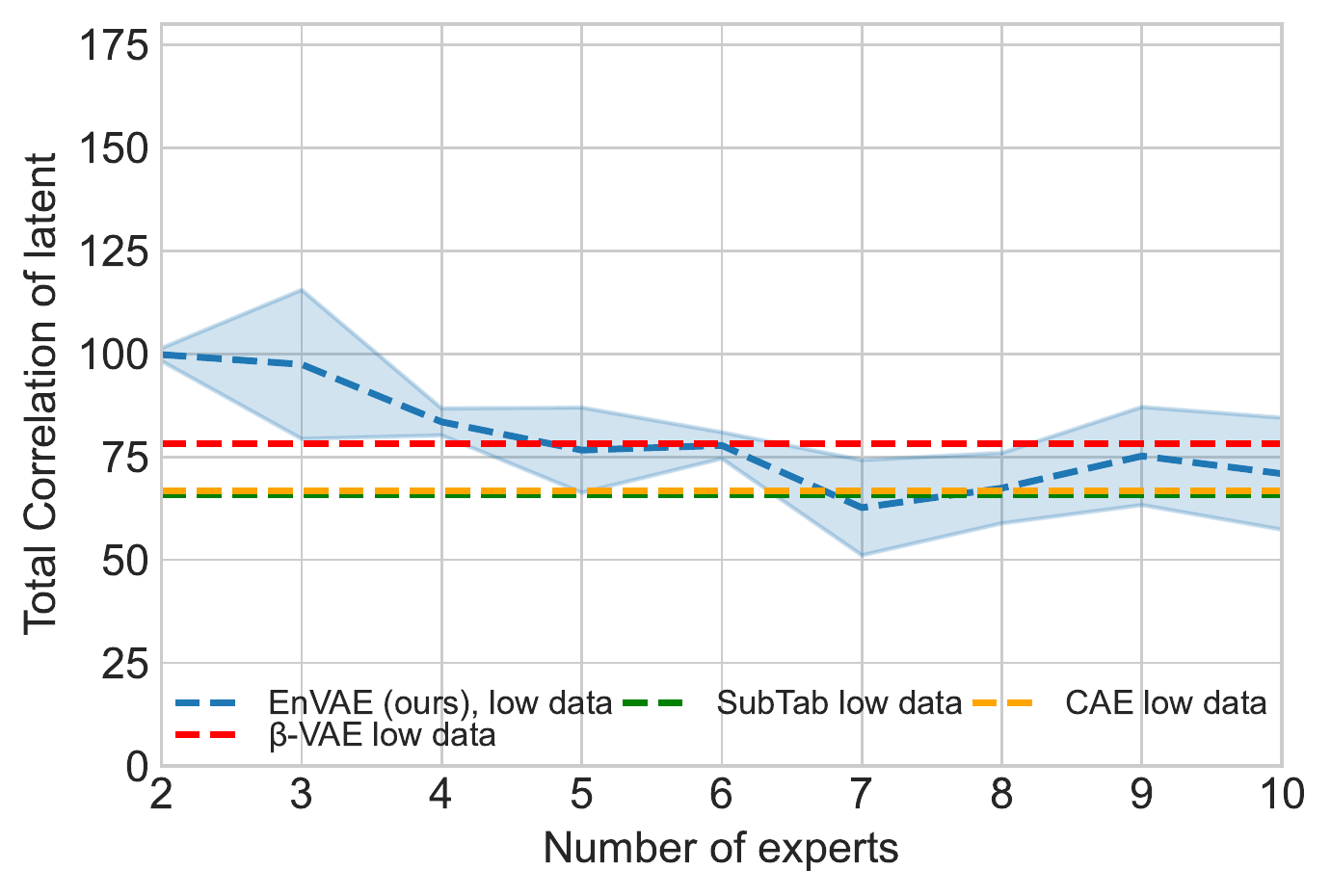}
        \vspace{-15pt}
    \caption{meta-pam}
\end{subfigure}
\begin{subfigure}[b]{0.49\textwidth}
    \includegraphics[width=\linewidth]{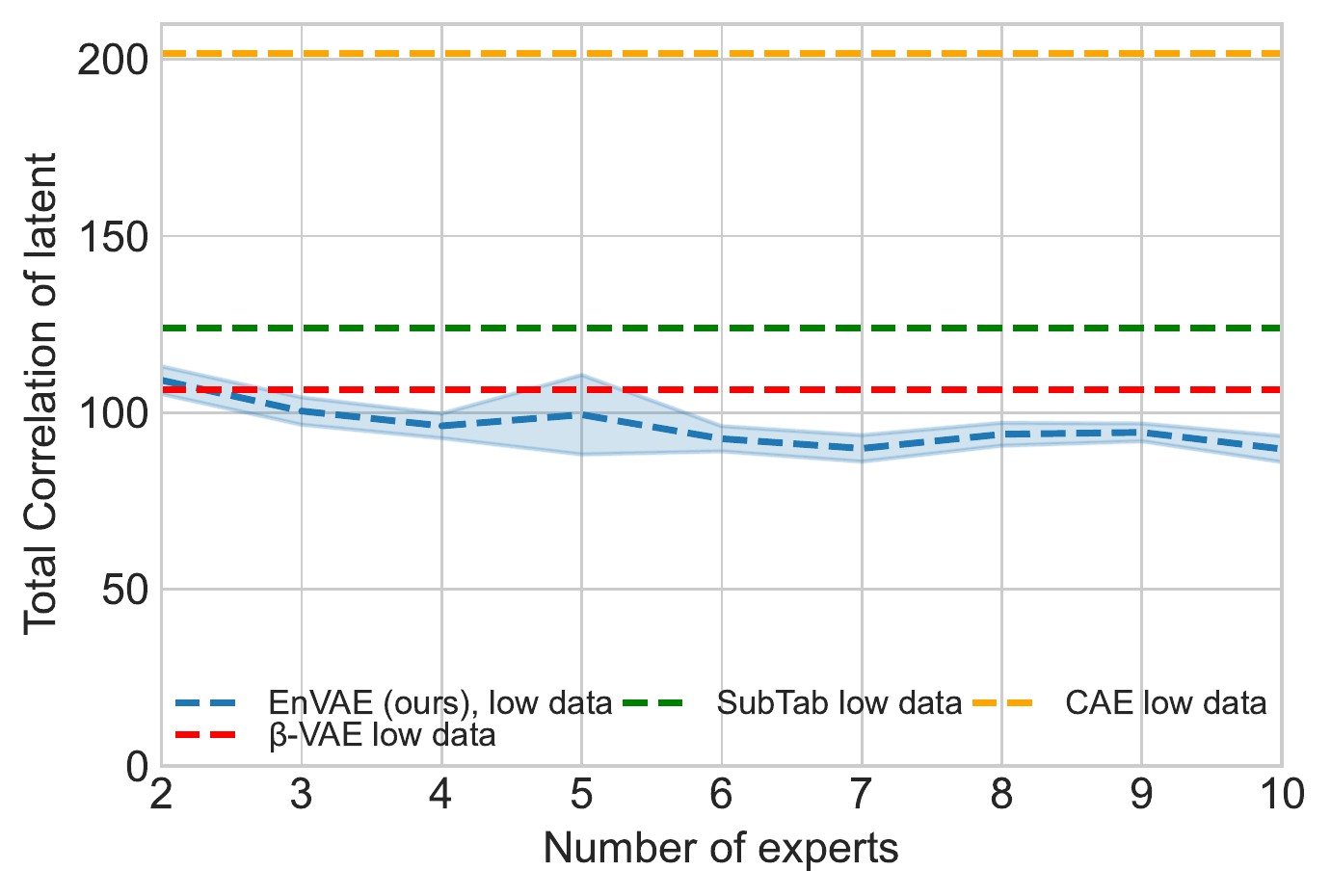}
        \vspace{-15pt}
    \caption{smk}
\end{subfigure}
\begin{subfigure}[b]{0.49\textwidth}
    \includegraphics[width=\linewidth]{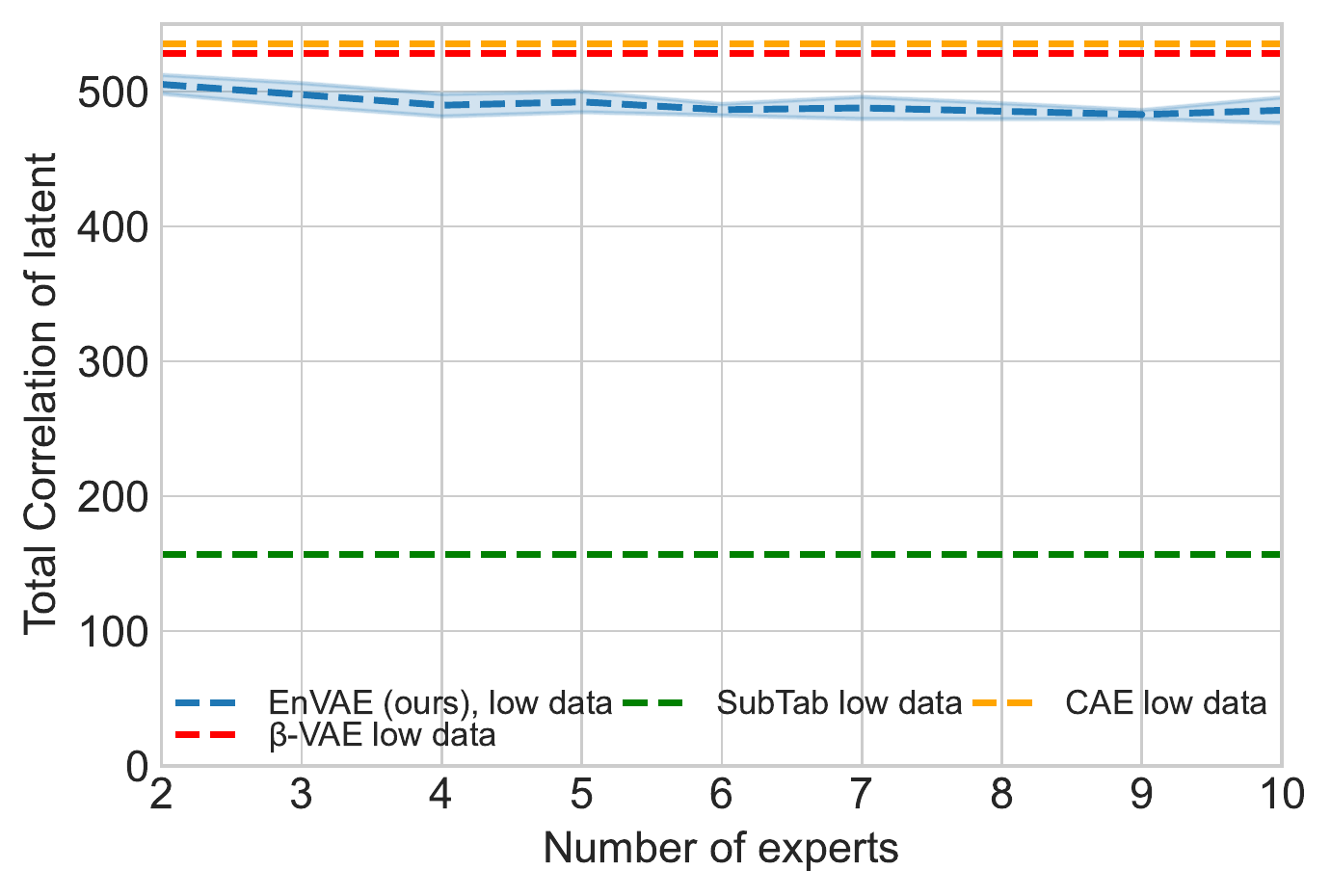}
        \vspace{-15pt}
    \caption{prostate}
\end{subfigure}
\begin{subfigure}[b]{0.49\textwidth}
    \includegraphics[width=\linewidth]{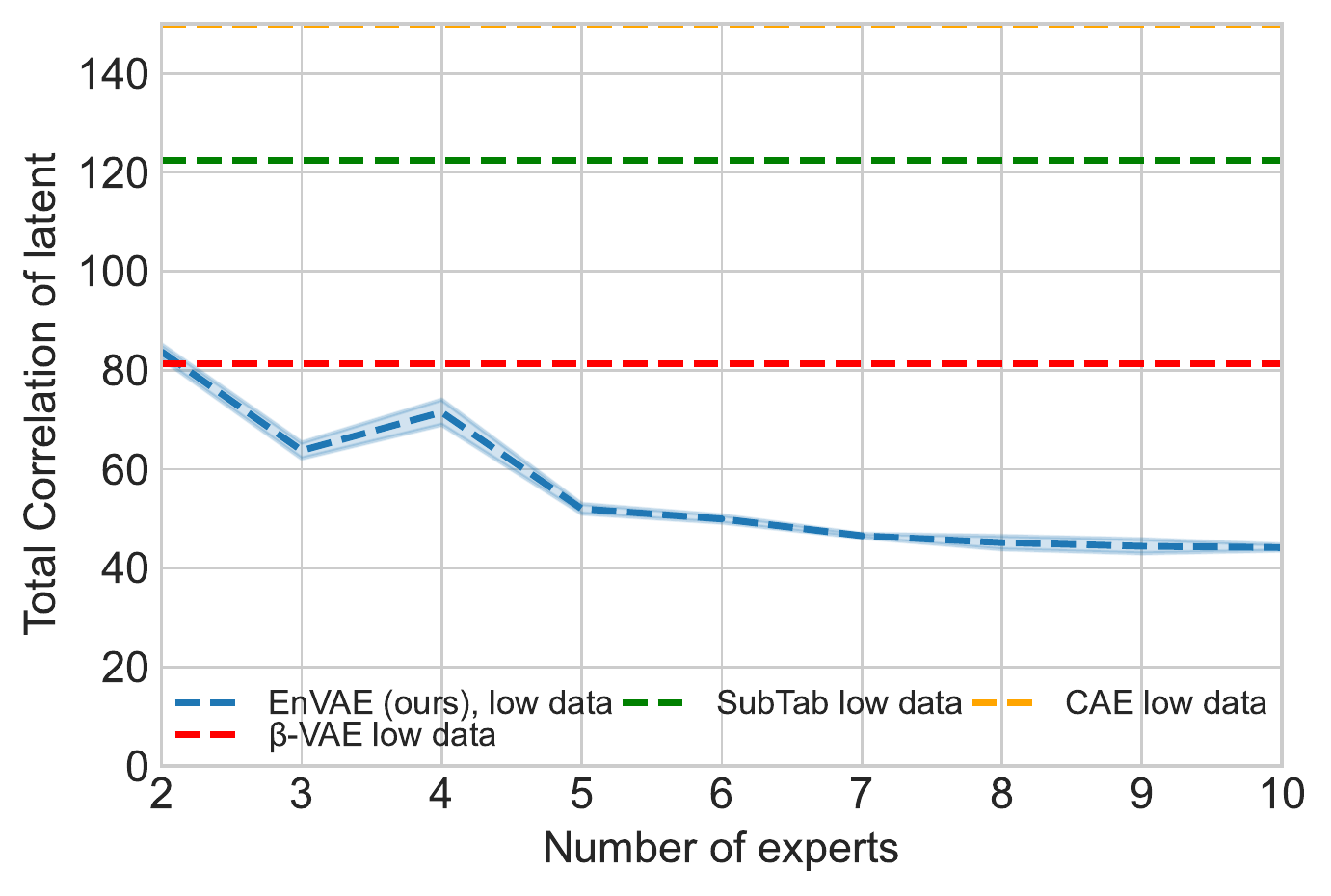}
        \vspace{-15pt}
    \caption{toxicity}
\end{subfigure}
\begin{subfigure}[b]{0.49\textwidth}
    \includegraphics[width=\linewidth]{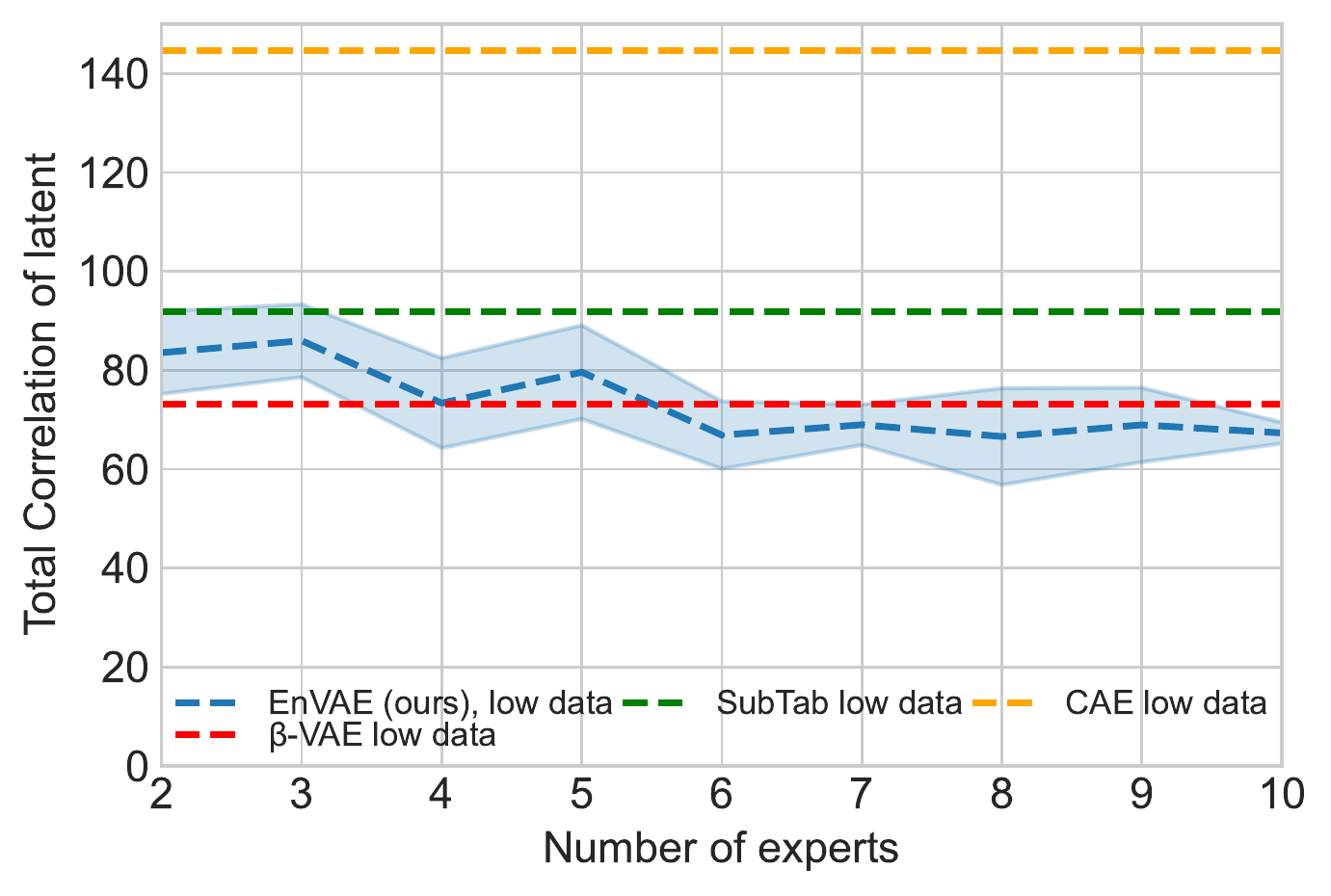}
        \vspace{-15pt}
    \caption{mpm}
\end{subfigure}
\begin{subfigure}[b]{0.49\textwidth}
    \includegraphics[width=\linewidth]{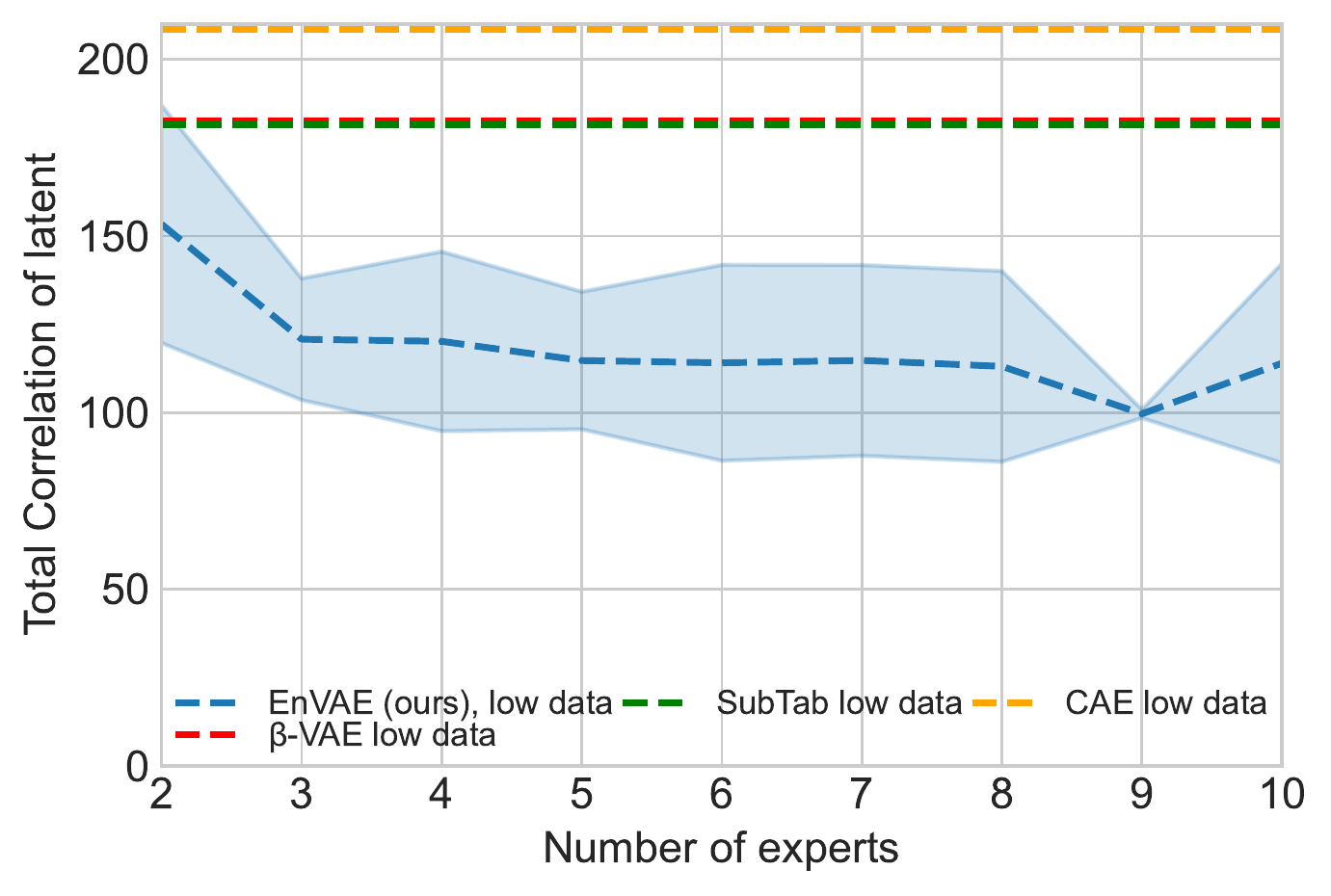}
        \vspace{-15pt}
    \caption{breast}
\end{subfigure}
\begin{subfigure}[b]{0.49\textwidth}
    \includegraphics[width=\linewidth]{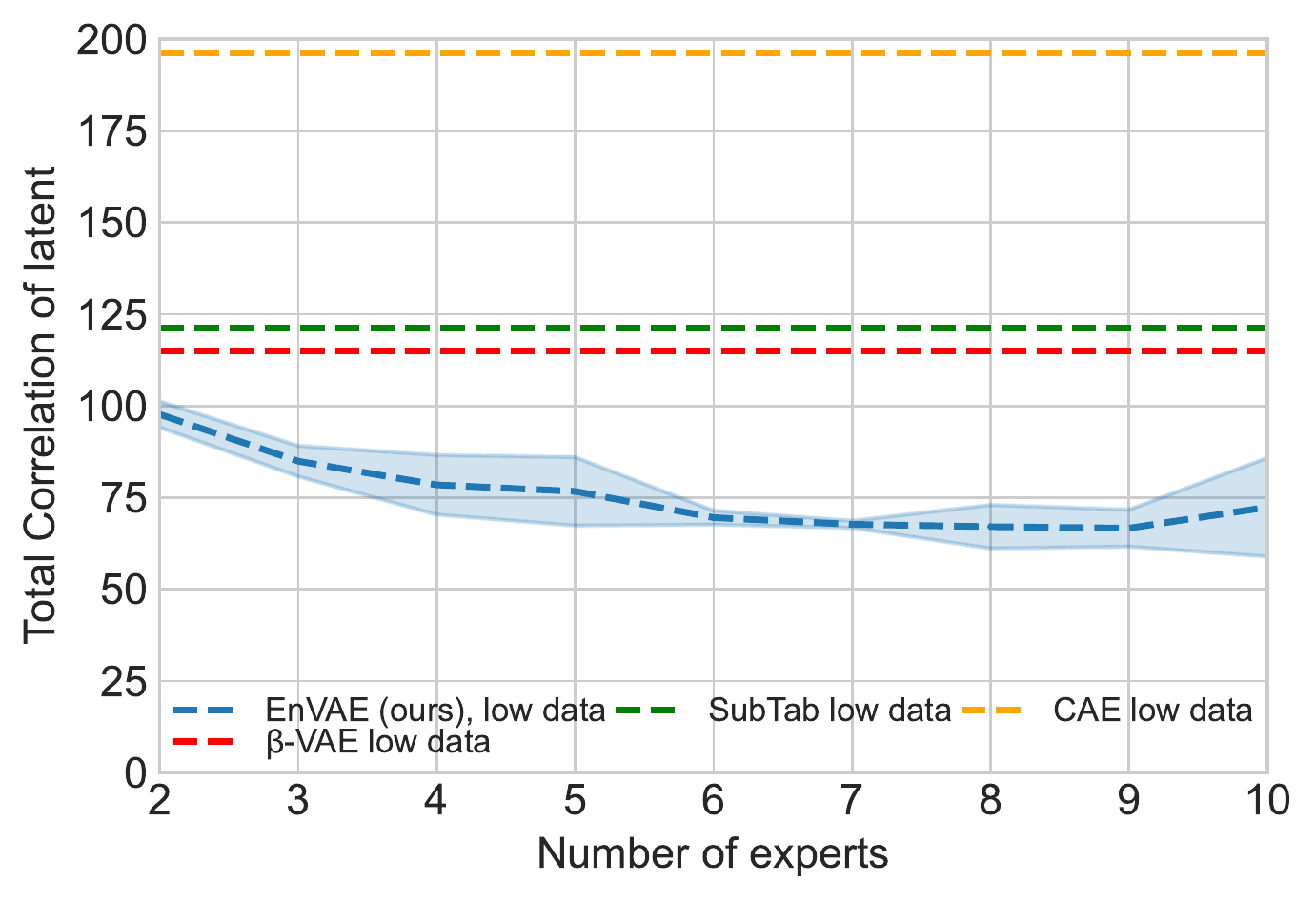}
        \vspace{-15pt}
    \caption{cll}
\end{subfigure}
    \caption{Total Correlation (TC) of the latent representations learnt on the \textit{smaller version of the datasets} by the benchmark models and EnVAE with varying numbers of experts. EnVAE consistently produces better latent representations, as indicated by a lower TC.}
    \label{fig:app_tc2}
\end{figure}

\newpage
\section{Visualisations of Latent Representations}\label{app:umap}
\begin{figure}[h!]
(1) `lung'\\
\begin{subfigure}[b]{0.24\textwidth}
         \centering
         \includegraphics[width=\textwidth]{images/umap/lung_CustomDataset_1_8_umap.pdf}
              \vspace{-16pt}
         % \caption{EnVAE (ours)}
     \end{subfigure}
     % \hfill
     \begin{subfigure}[b]{0.24\textwidth}
         \centering
         \includegraphics[width=\textwidth]{images/umap/beta_vae_lung_1_0.125_umap.pdf}
              \vspace{-16pt}
         % \caption{$\beta$-VAE}
     \end{subfigure}
      \begin{subfigure}[b]{0.24\textwidth}
         \centering
         \includegraphics[width=\textwidth]{images/umap/cae_lung_1_300_umap.pdf}
              \vspace{-16pt}
     %     \caption{CAE}
     \end{subfigure}
      \begin{subfigure}[b]{0.24\textwidth}
         \centering
         \includegraphics[width=\textwidth]{images/umap/subtab_lung_1_4_0.75_umap.pdf}
      \vspace{-16pt}
         % \caption{SubTab}
\end{subfigure}
(2) `toxicity'\\

\begin{subfigure}[b]{0.24\textwidth}
         \centering
         \includegraphics[width=\textwidth]{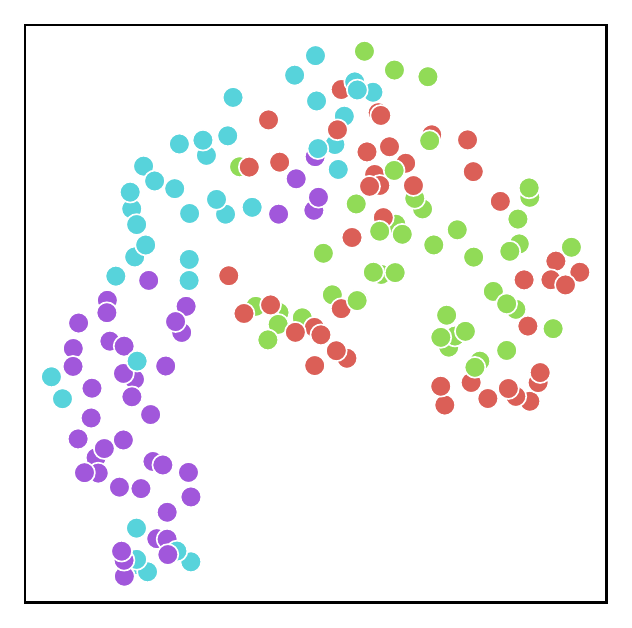}
              \vspace{-16pt}
         % \caption{EnVAE (ours)}
     \end{subfigure}
     % \hfill
     \begin{subfigure}[b]{0.24\textwidth}
         \centering
         \includegraphics[width=\textwidth]{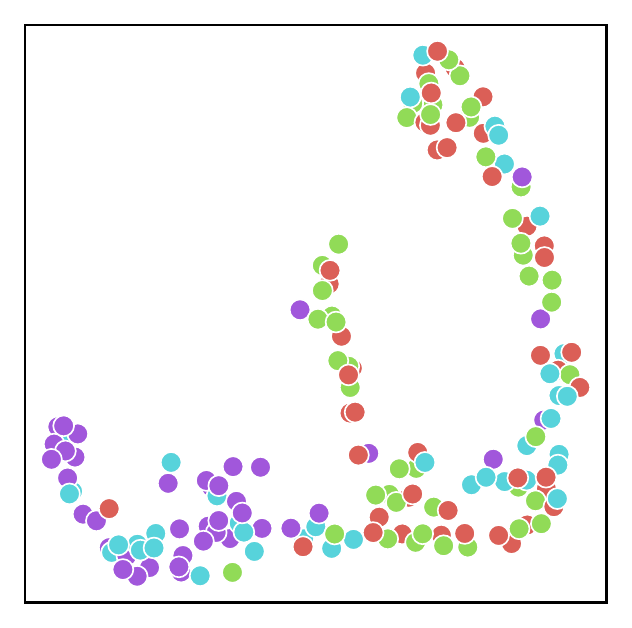}
              \vspace{-16pt}
         % \caption{$\beta$-VAE}
     \end{subfigure}
      \begin{subfigure}[b]{0.24\textwidth}
         \centering
         \includegraphics[width=\textwidth]{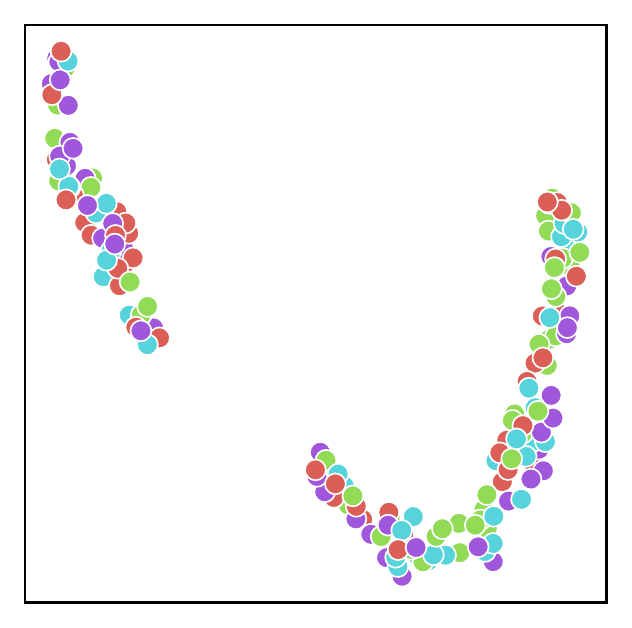}
              \vspace{-16pt}
     %     \caption{CAE}
     \end{subfigure}
      \begin{subfigure}[b]{0.24\textwidth}
         \centering
         \includegraphics[width=\textwidth]{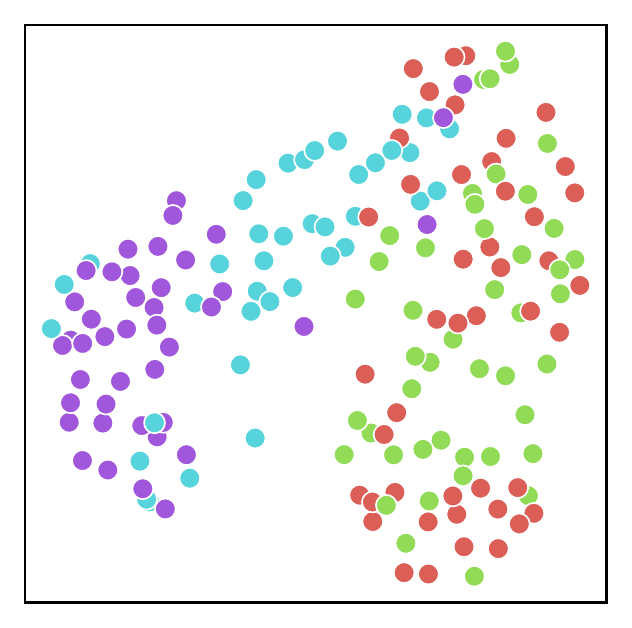}
      \vspace{-16pt}
         % \caption{SubTab}
\end{subfigure}
(3) `mpm'\\

\begin{subfigure}[b]{0.24\textwidth}
         \centering
         \includegraphics[width=\textwidth]{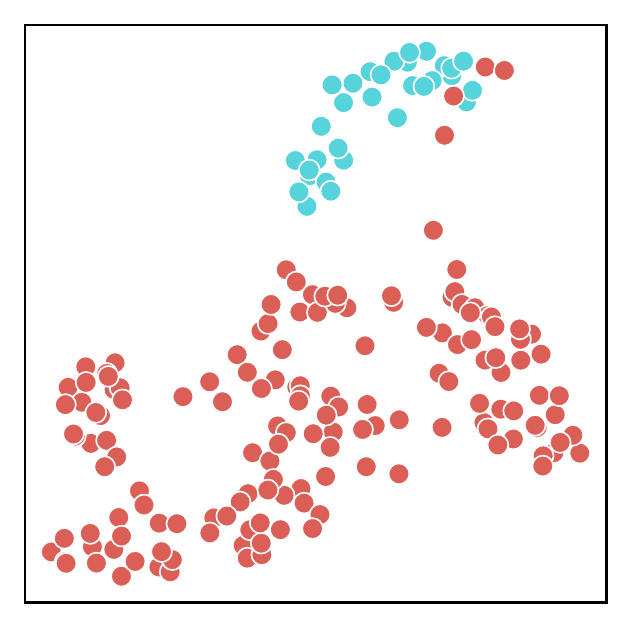}
              \vspace{-16pt}
         % \caption{EnVAE (ours)}
     \end{subfigure}
     % \hfill
     \begin{subfigure}[b]{0.24\textwidth}
         \centering
         \includegraphics[width=\textwidth]{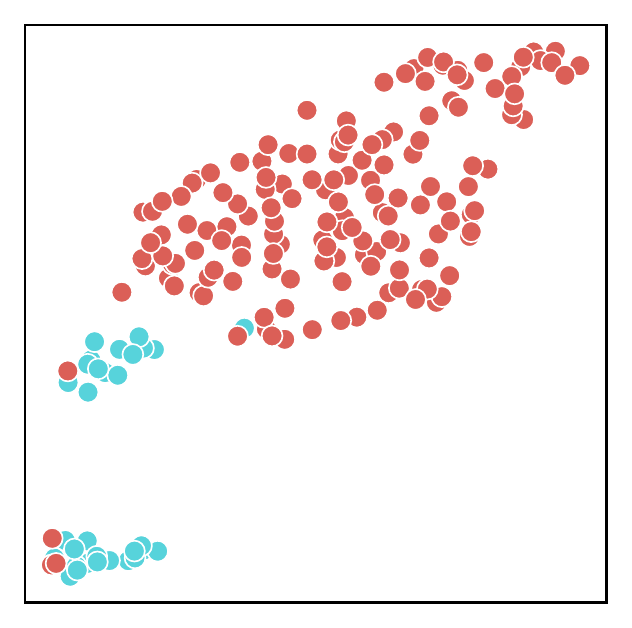}
              \vspace{-16pt}
         % \caption{$\beta$-VAE}
     \end{subfigure}
      \begin{subfigure}[b]{0.24\textwidth}
         \centering
         \includegraphics[width=\textwidth]{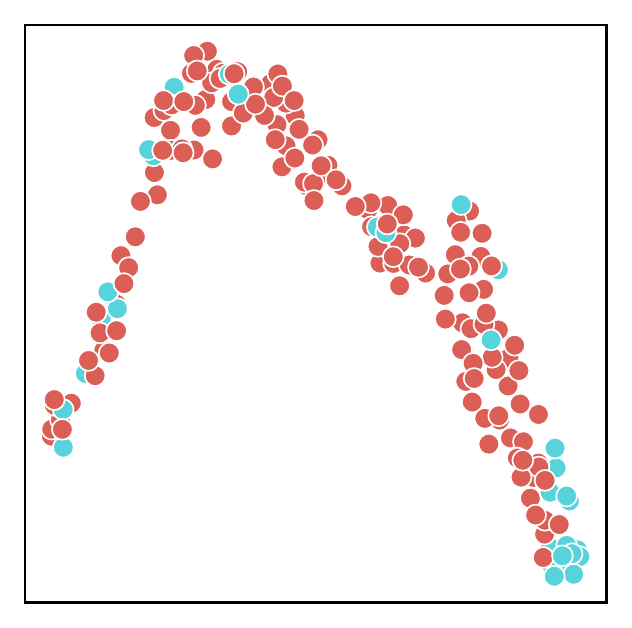}
              \vspace{-16pt}
     %     \caption{CAE}
     \end{subfigure}
      \begin{subfigure}[b]{0.24\textwidth}
         \centering
         \includegraphics[width=\textwidth]{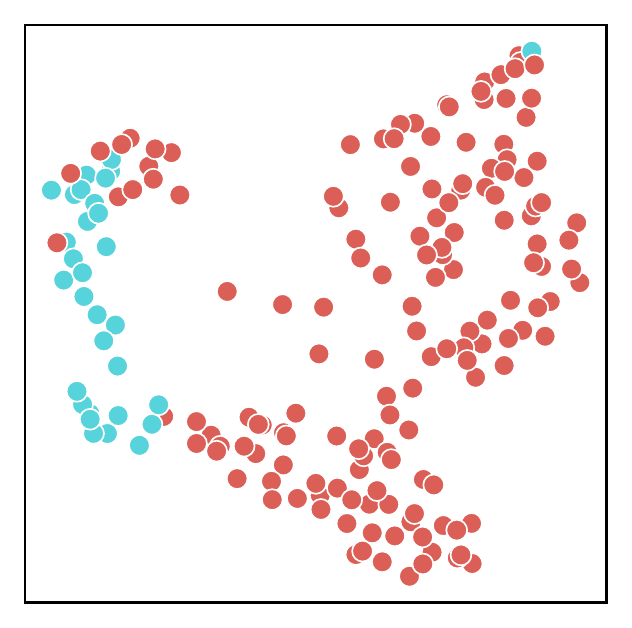}
      \vspace{-16pt}
         % \caption{SubTab}
\end{subfigure}

(4) `meta-pam'\\
\begin{subfigure}[b]{0.24\textwidth}
         \centering
         \includegraphics[width=\textwidth]{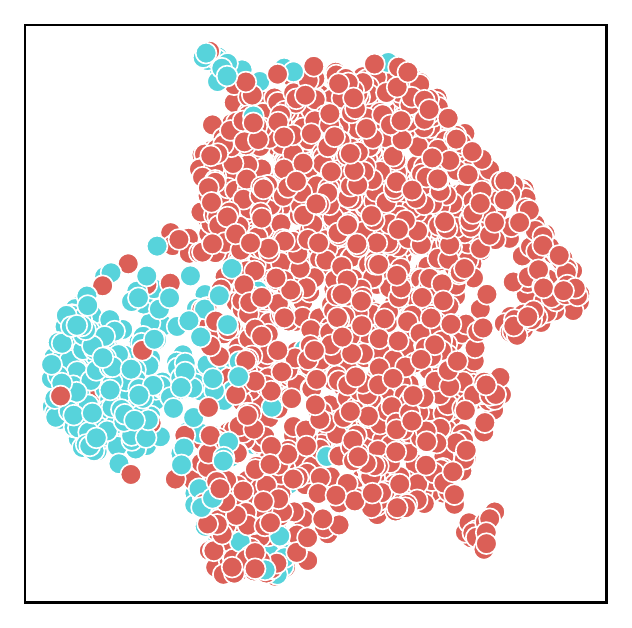}
              \vspace{-16pt}
         \caption{EnVAE (ours)}
     \end{subfigure}
     % \hfill
     \begin{subfigure}[b]{0.24\textwidth}
         \centering
         \includegraphics[width=\textwidth]{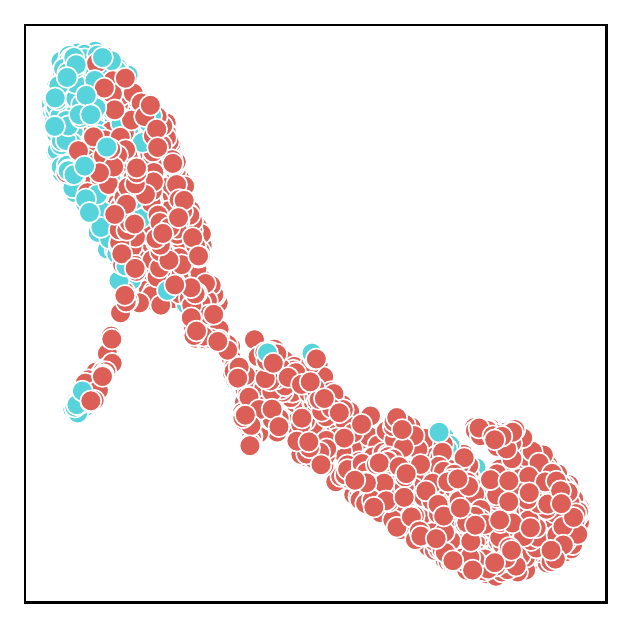}
              \vspace{-16pt}
         \caption{$\beta$-VAE}
     \end{subfigure}
      \begin{subfigure}[b]{0.24\textwidth}
         \centering
         \includegraphics[width=\textwidth]{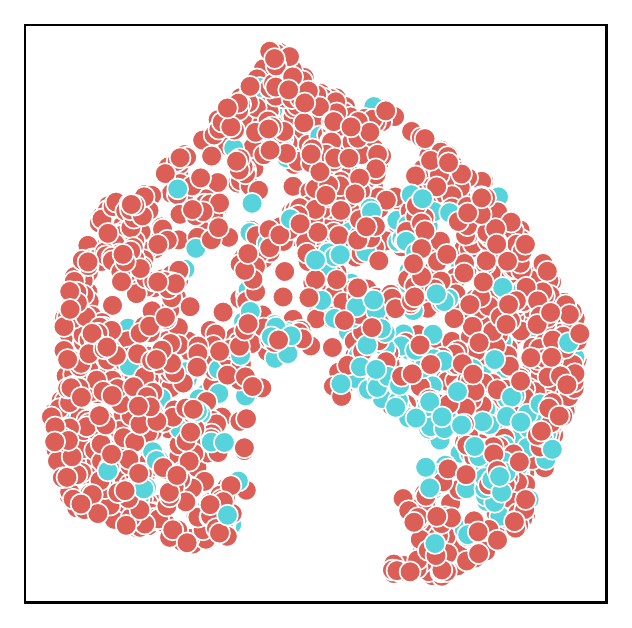}
              \vspace{-16pt}
         \caption{CAE}
     \end{subfigure}
      \begin{subfigure}[b]{0.24\textwidth}
         \centering
         \includegraphics[width=\textwidth]{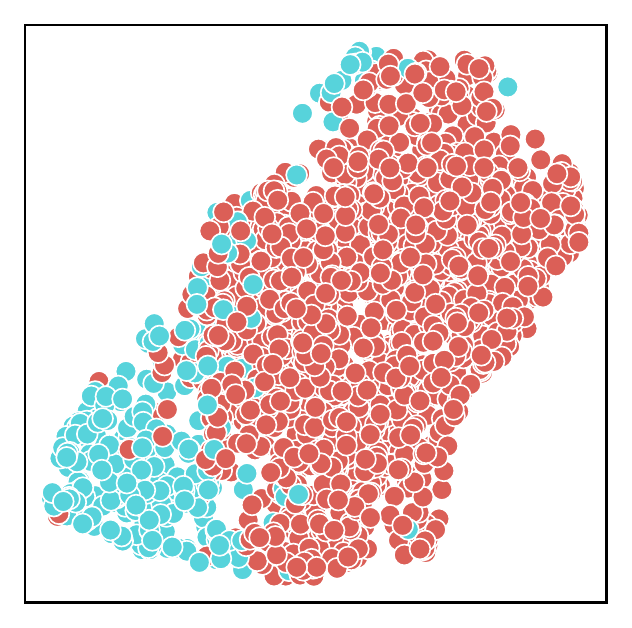}
      \vspace{-16pt}
         \caption{SubTab}
     \end{subfigure}

      \caption{EnVAE can be used for data exploration. We plot using UMAP the latent representations on \textbf{(top)} `lung', \textbf{(middle-top)} `toxicity'  , \textbf{(middle-bottom)} `mpm'  and \textbf{(bottom)} meta-pam, where each point represents the embedding of one sample, and the colour represents its label. The representations learned using EnVAE show that the samples cluster comparatively better to the other techniques.}
    \label{fig:umap1}
\end{figure}

\begin{figure}[h!]
(5) `cll'\\
\begin{subfigure}[b]{0.24\textwidth}
         \centering
         \includegraphics[width=\textwidth]{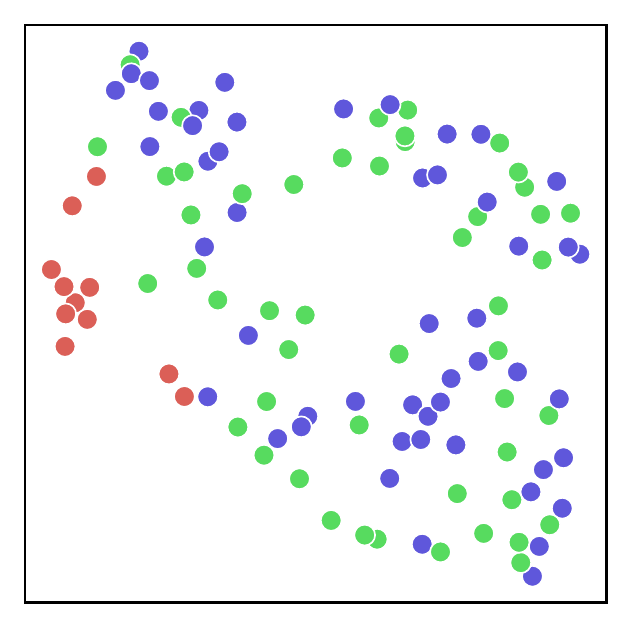}
              \vspace{-16pt}
         \caption{EnVAE (ours)}
     \end{subfigure}
     % \hfill
     \begin{subfigure}[b]{0.24\textwidth}
         \centering
         \includegraphics[width=\textwidth]{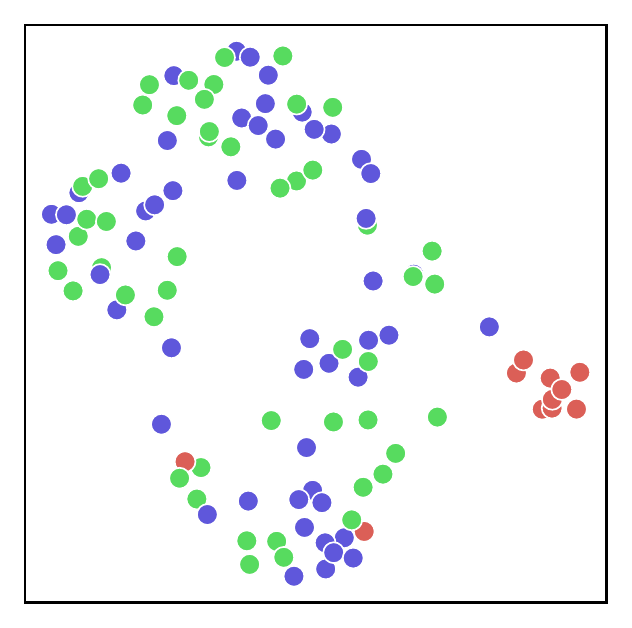}
              \vspace{-16pt}
         \caption{$\beta$-VAE}
     \end{subfigure}
      \begin{subfigure}[b]{0.24\textwidth}
         \centering
         \includegraphics[width=\textwidth]{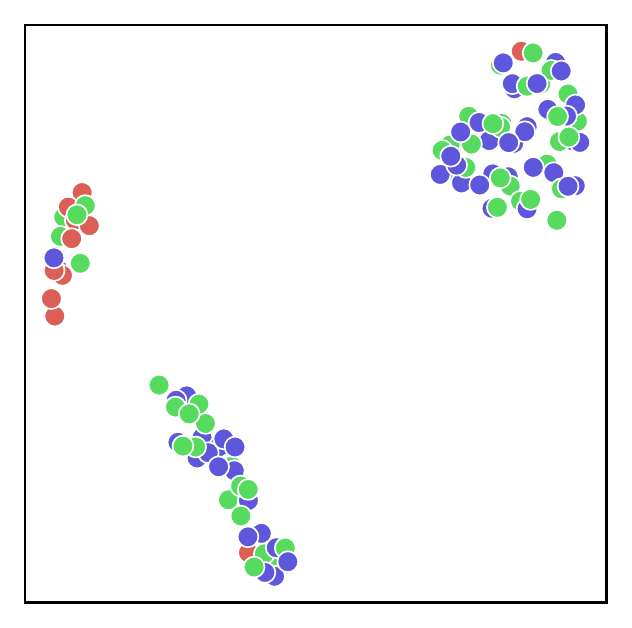}
              \vspace{-16pt}
         \caption{CAE}
     \end{subfigure}
      \begin{subfigure}[b]{0.24\textwidth}
         \centering
         \includegraphics[width=\textwidth]{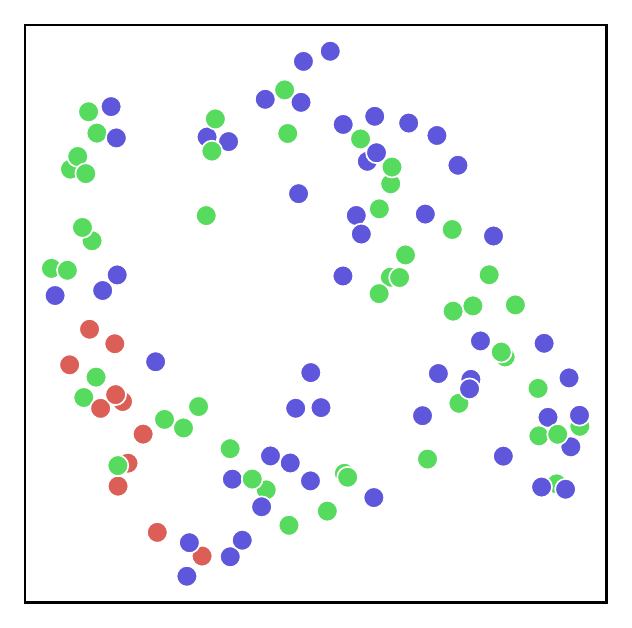}
      \vspace{-16pt}
         \caption{SubTab}
     \end{subfigure}
(6) `smk'\\
\begin{subfigure}[b]{0.24\textwidth}
         \centering
         \includegraphics[width=\textwidth]{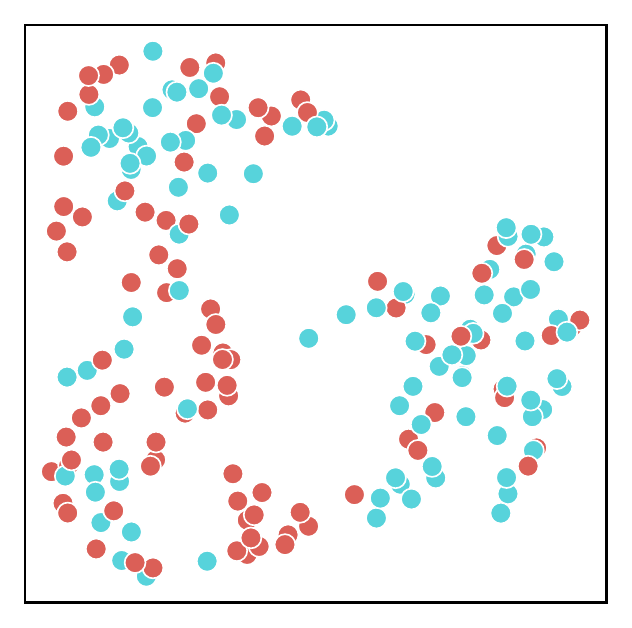}
              \vspace{-16pt}
         \caption{EnVAE (ours)}
     \end{subfigure}
     % \hfill
     \begin{subfigure}[b]{0.24\textwidth}
         \centering
         \includegraphics[width=\textwidth]{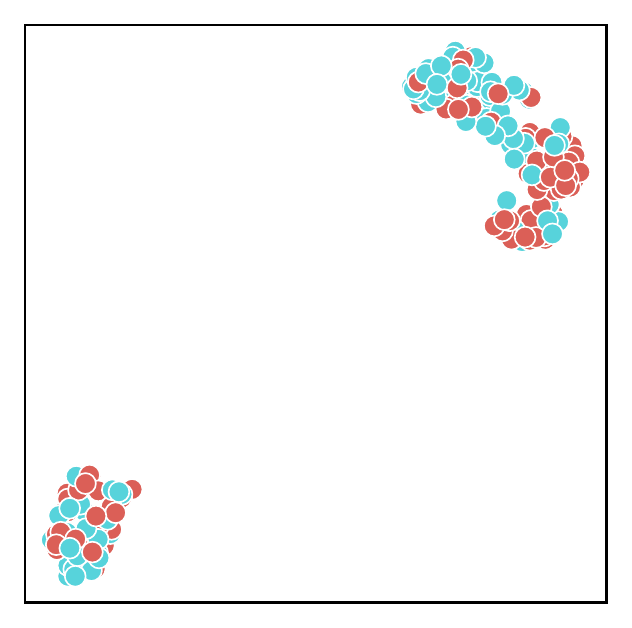}
              \vspace{-16pt}
         \caption{$\beta$-VAE}
     \end{subfigure}
      \begin{subfigure}[b]{0.24\textwidth}
         \centering
         \includegraphics[width=\textwidth]{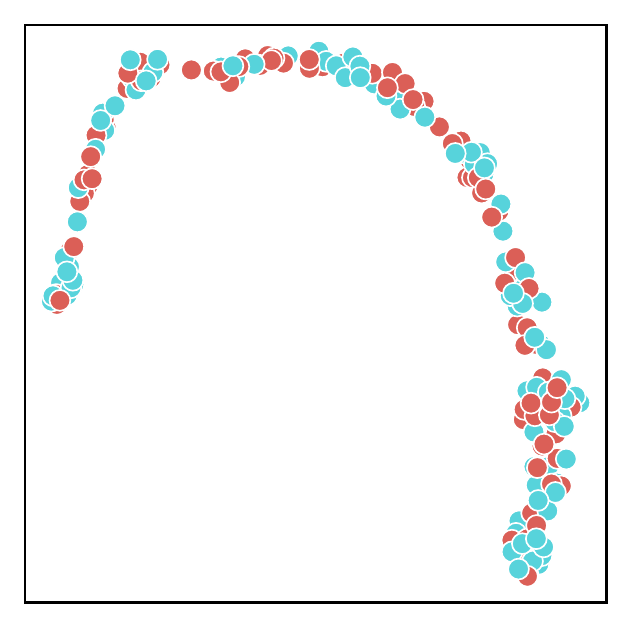}
              \vspace{-16pt}
         \caption{CAE}
     \end{subfigure}
      \begin{subfigure}[b]{0.24\textwidth}
         \centering
         \includegraphics[width=\textwidth]{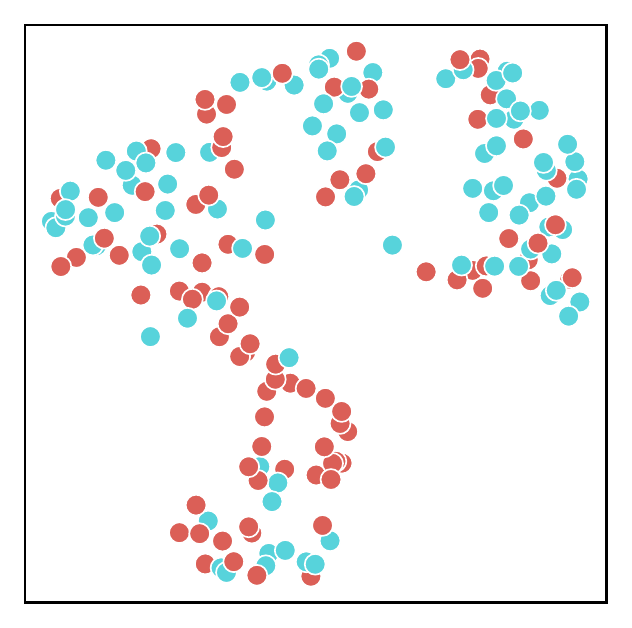}
      \vspace{-16pt}
         \caption{SubTab}
     \end{subfigure}
(7) `breast'\\
\begin{subfigure}[b]{0.24\textwidth}
         \centering
         \includegraphics[width=\textwidth]{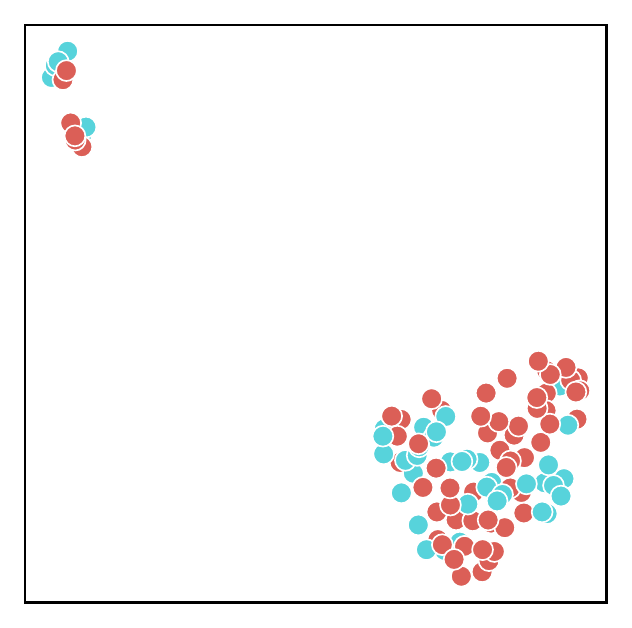}
              \vspace{-16pt}
         \caption{EnVAE (ours)}
     \end{subfigure}
     % \hfill
     \begin{subfigure}[b]{0.24\textwidth}
         \centering
         \includegraphics[width=\textwidth]{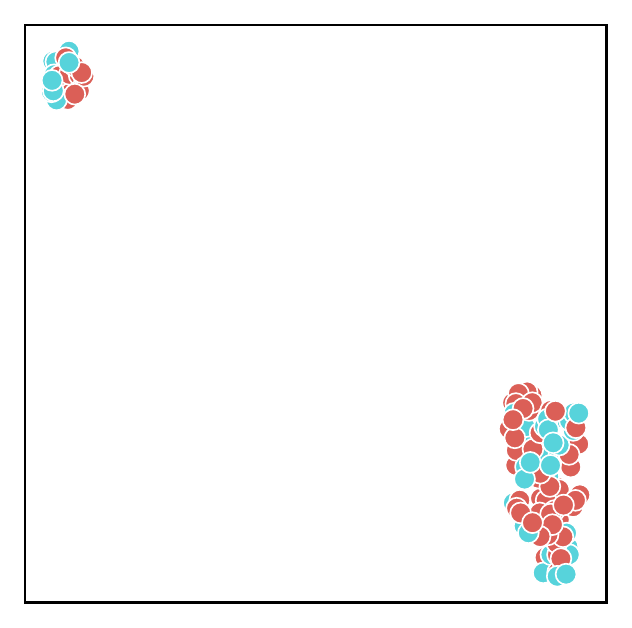}
              \vspace{-16pt}
         \caption{$\beta$-VAE}
     \end{subfigure}
      \begin{subfigure}[b]{0.24\textwidth}
         \centering
         \includegraphics[width=\textwidth]{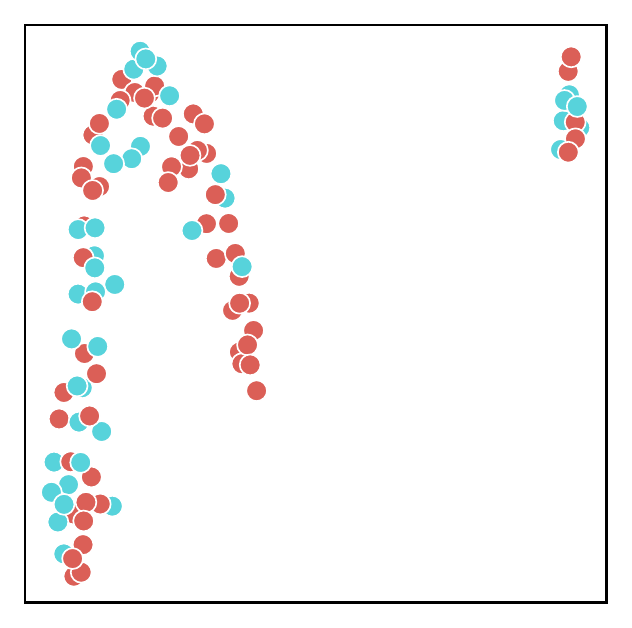}
              \vspace{-16pt}
         \caption{CAE}
     \end{subfigure}
      \begin{subfigure}[b]{0.24\textwidth}
         \centering
         \includegraphics[width=\textwidth]{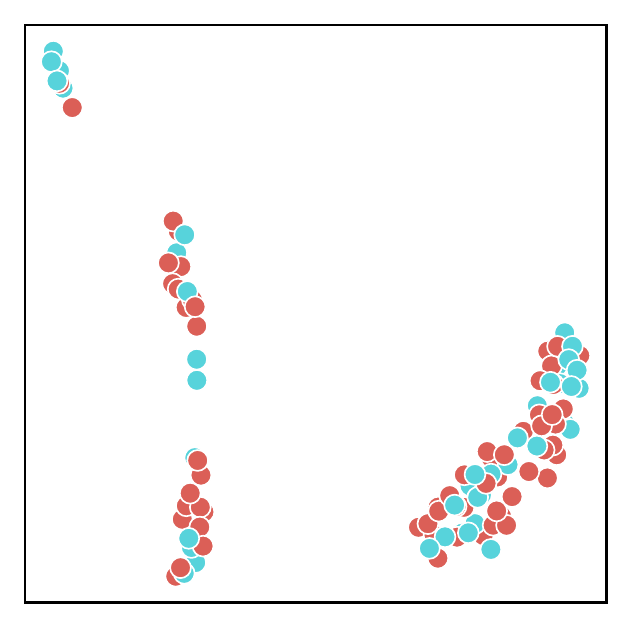}
      \vspace{-16pt}
         \caption{SubTab}
     \end{subfigure}
(8) `prostate'\\
\begin{subfigure}[b]{0.24\textwidth}
         \centering
         \includegraphics[width=\textwidth]{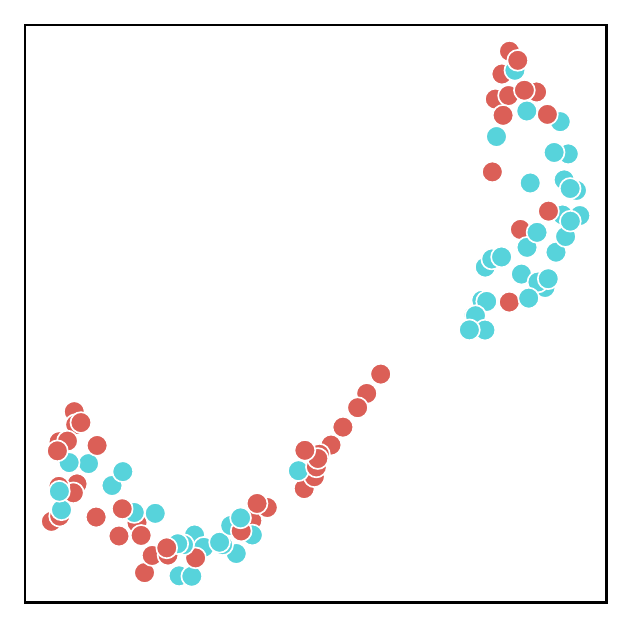}
              \vspace{-16pt}
         \caption{EnVAE (ours)}
     \end{subfigure}
     % \hfill
     \begin{subfigure}[b]{0.24\textwidth}
         \centering
         \includegraphics[width=\textwidth]{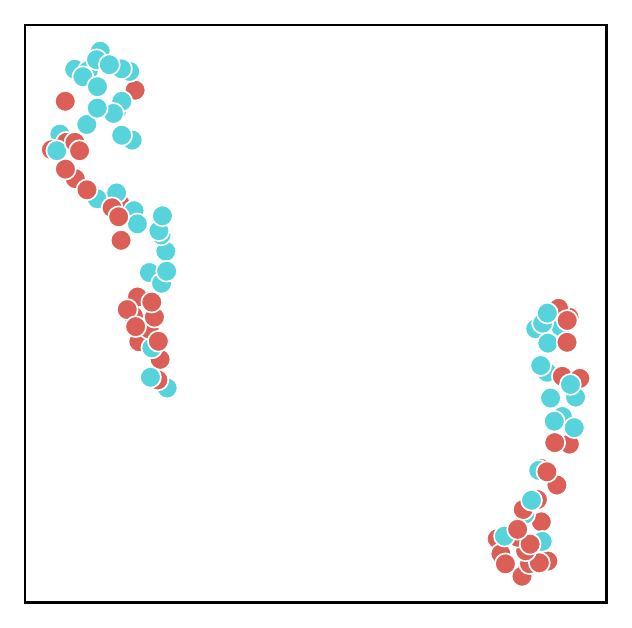}
              \vspace{-16pt}
         \caption{$\beta$-VAE}
     \end{subfigure}
      \begin{subfigure}[b]{0.24\textwidth}
         \centering
         \includegraphics[width=\textwidth]{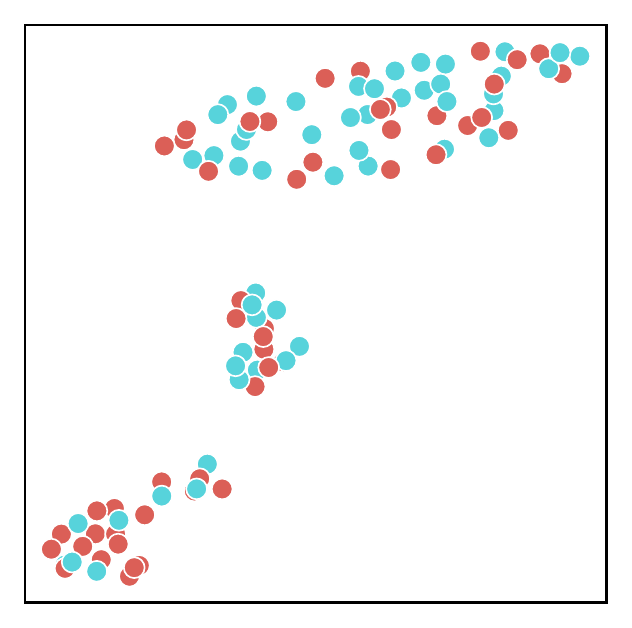}
              \vspace{-16pt}
         \caption{CAE}
     \end{subfigure}
      \begin{subfigure}[b]{0.24\textwidth}
         \centering
         \includegraphics[width=\textwidth]{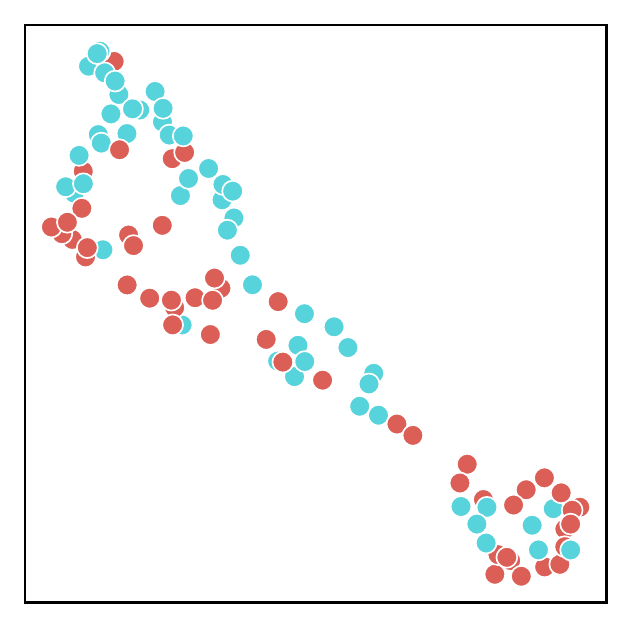}
      \vspace{-16pt}
         \caption{SubTab}
     \end{subfigure}
         % \vspace{-3pt}
     \caption{EnVAE can be used for data exploration. We plot using UMAP the latent representations on \textbf{(top)} `cll', \textbf{(middle-top)} `smk'  , \textbf{(middle-bottom)} `breast'  and \textbf{(bottom)} `prostate', where each point represents the embedding of one sample, and the colour represents its label. The representations learned using EnVAE show that the samples cluster comparatively better to the other techniques.}
    \label{fig:umap2}
\end{figure}

We present the latent representations learnt by all models across datasets using UMAP~\cite{mcinnes_umap_2018} with 15 neighbours, a minimum distance of 0.3 and using the cosine metric and 2 components. We transform and visualise the entire dataset as the test set has too few samples to effectively visualise. Figures \ref{fig:umap1} and \ref{fig:umap2} visualise the clusters learnt by each model. 

\vfill\clearpage

\section{EnVAE Algorithm}
\begin{algorithm}
\caption{Training EnVAE}\label{alg:envae}
\begin{algorithmic}
\State \textbf{Input:} training data $\mathcal{X}\in \mathbb{R}^{N\times D}$, encoder networks $Enc_1,...,Enc_m$ with parameters $\phi$ and decoder networks $Dec_1,...,Dec_m$ with parameters $\theta$ and learning rate $\alpha$
\For{each batch $B= \{\pmb{x}^{(i)}\}^b_{i=1}$}
    \For{each group, $\pmb{x}_i$}
        \State{$q_\phi(\pmb{z}|\pmb{x_i}) = Enc_i(\pmb{x}_i)$}
    \EndFor
    \State $q_\phi(\pmb{z}|\pmb{x}) = 0$
    \For{$A \in \mathcal{P}(N_m)$}
        \State $q_\phi(\pmb{z}| \pmb{x}_A) = \prod_{i \in A} q_\theta(\pmb{z}| \pmb{x}_i)$
        \State $q_\phi(\pmb{z}|\pmb{x}) \mathrel{+}= \frac{1}{|\mathcal{P}(N_m)|} q_\phi(\pmb{z}| \pmb{x}_A)$
    \EndFor
    \State{Sample $q_\phi(\pmb{z}|\pmb{x})$ to obtain $\pmb{z}$}
    \For{each group, $\pmb{x}_i$}
        \State $\hat{\pmb{x}}_i = Dec_i(\pmb{z})$
    \EndFor
    \State{Compute the training loss $L$:}
    \Indent
        \State{$L = MeanSquareError(\texttt{concat}(\hat{\pmb{x}}_1,...,\hat{\pmb{x}}_m), \pmb{x}) + KL(q_\phi(\pmb{z}|\pmb{x}) || p(\pmb{z}))$}
    \EndIndent
    \State{Compute the gradient of $L$ w.r.t:}
    \Indent
        \State{$\phi, \theta$ using backpropagation}
    \EndIndent
    \State{Update the parameters}
    \Indent
        \State{$\phi \leftarrow \phi-\alpha \nabla_\phi L$}
        \State{$\theta \leftarrow \theta-\alpha \nabla_\theta L$}
    \EndIndent        
\EndFor
\end{algorithmic}
\end{algorithm}

\end{document}